\documentclass[10pt,journal,compsoc]{IEEEtran}
\usepackage{graphicx}
\usepackage{booktabs}
\usepackage{subcaption}
\usepackage{epstopdf}
\usepackage{diagbox}
\usepackage{threeparttable}
\usepackage{makecell,multirow,diagbox}

\usepackage[linesnumbered,ruled,vlined]{algorithm2e}
\usepackage{algpseudocode}
\usepackage{amssymb}
\usepackage{amsmath}
\usepackage{amsthm}
\usepackage{cite}
\usepackage{hyperref}
\usepackage{bm}
\usepackage{mathtools}
\usepackage{pgfmath}
\usepackage{dsfont}
\usepackage{xcolor}
\usepackage{color}
\usepackage{lineno}

\usepackage{flushend}

\def\ourmethod{TFormer}

\def\BibTeX{{\rm B\kern-.05em{\sc i\kern-.025em b}\kern-.08em
    T\kern-.1667em\lower.7ex\hbox{E}\kern-.125emX}}

\begin{document}
\title{\ourmethod{}: A Transmission-Friendly ViT Model\\for IoT Devices}

\author{Zhichao~Lu,
        ~Chuntao~Ding,
        ~Felix~Juefei-Xu,~\IEEEmembership{Member,~IEEE,}
        ~Vishnu~Naresh~Boddeti,~\IEEEmembership{Member,~IEEE,} \\
        ~Shangguang~Wang,~\IEEEmembership{Senior Member,~IEEE,}
        ~Yun~Yang,~\IEEEmembership{Senior Member,~IEEE,}
\IEEEcompsocitemizethanks{\IEEEcompsocthanksitem Zhichao Lu is with the School of Software Engineering, Sun Yat-sen University, Zhuhai 519082, China. E-mail: luzhichaocn@gmail.com.
\IEEEcompsocthanksitem Chuntao Ding is with the School of Computer and Information Technology, Beijing Jiaotong University, Beijing, China.
E-mail: chtding@bjtu.edu.cn.
\IEEEcompsocthanksitem Felix Juefei-Xu is with Meta AI, New York, NY 10001 and New York University, New York, NY 10012, USA. This work is done prior to joining Meta AI. 
                E-mail: felixu@meta.com, juefei.xu@nyu.edu.
\IEEEcompsocthanksitem Vishnu N. Boddeti is with the Department of Computer Science and Engineering,
                Michigan State University, East Lansing, MI 48824, USA.
                E-mail: vishnu@msu.edu.
\IEEEcompsocthanksitem Shangguang Wang is with the Key Laboratory of Networking and Switching Technology, Beijing University of Posts and Telecommunications, Beijing, China 100876.
E-mail: {sgwang@bupt.edu.cn}.
\IEEEcompsocthanksitem Yun Yang is with the Department of Computing Technologies, Swinburne University of Technology, Melbourne, Australia.
E-mail: {yyang@swin.edu.au}.

(Corresponding author: Chuntao Ding.)
}
}

\markboth{IEEE Transactions on Parallel and Distributed Systems}%
{Shell \MakeLowercase{\textit{et al.}}: Bare Demo of IEEEtran.cls for IEEE Transactions on Magnetics Journals}

\IEEEtitleabstractindextext{%
\begin{abstract}
Deploying high-performance vision transformer (ViT) models on ubiquitous Internet of Things (IoT) devices to provide high-quality vision services will revolutionize the way we live, work, and interact with the world.
Due to the contradiction between the limited resources of IoT devices and resource-intensive ViT models, the use of cloud servers to assist ViT model training has become mainstream. 
However, due to the larger number of parameters and floating-point operations (FLOPs) of the existing ViT models, the model parameters transmitted by cloud servers are large and difficult to run on resource-constrained IoT devices.
To this end, this paper proposes a transmission-friendly ViT model, TFormer, for deployment on resource-constrained IoT devices with the assistance of a cloud server.
The high performance and small number of model parameters and FLOPs of TFormer are attributed to the proposed hybrid layer and the proposed partially connected feed-forward network (PCS-FFN).
The hybrid layer consists of nonlearnable modules and a pointwise convolution, which can obtain multitype and multiscale features with only a few parameters and FLOPs to improve the TFormer performance.
The PCS-FFN adopts group convolution to reduce the number of parameters.
The key idea of this paper is to propose TFormer with few model parameters and FLOPs to facilitate applications running on resource-constrained IoT devices to benefit from the high performance of the ViT models.
Experimental results on the ImageNet-1K, MS COCO, and ADE20K datasets for image classification, object detection, and semantic segmentation tasks demonstrate that the proposed model outperforms other state-of-the-art models. 
Specifically, \ourmethod{}-S achieves 5\% higher accuracy on ImageNet-1K than ResNet18 with 1.4$\times$ fewer parameters and FLOPs. 
\end{abstract}
}

\maketitle
\IEEEpeerreviewmaketitle

\section{Introduction} \label{ref-introduction}
The International Data Corporation predicts that by 2025, there will be 41.6 billion connected Internet of Things (IoT) devices \cite{IDCPredicted}.
Additionally, the recently proposed vision transformer (ViT) models, with the support of large datasets, have crushed the convolutional neural network models that have dominated for many years in multifarious vision tasks, such as image classification~\cite{Dosovitskiy@An,Touvron@Training}, object detection~\cite{Carion@End,Liu@Swin}, and semantic segmentation~\cite{Zheng@Rethinking,Strudel@Segmenter}.  
Deploying high-performance ViT models on ubiquitous IoT devices to provide high-quality vision services has attracted great attention from both industry and academia.
\begin {figure}[t]
\centering
\includegraphics[width=.95\linewidth]{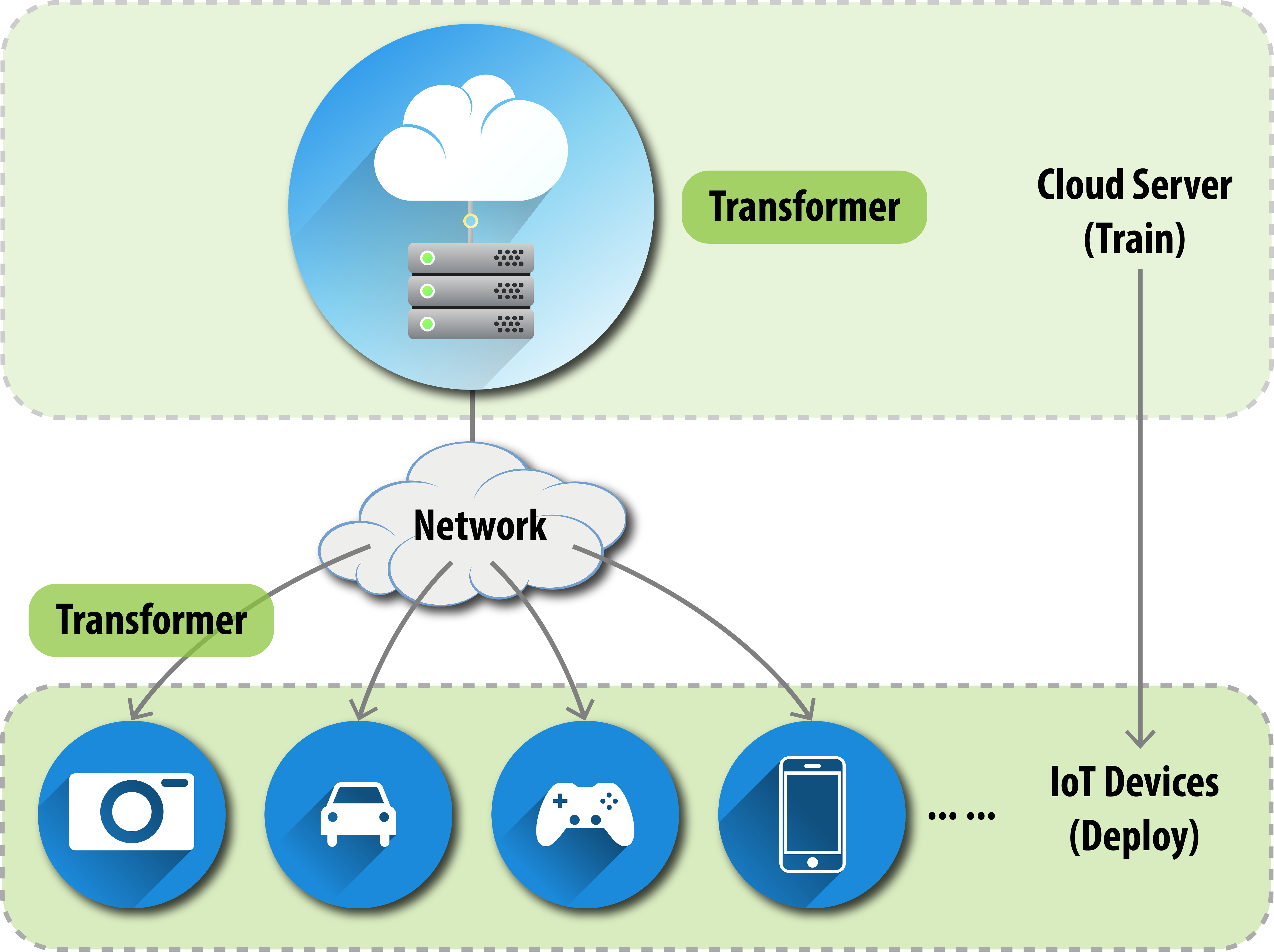}
\vspace{-12pt}
\caption {Overview of system architecture. \vspace{-3em}}
\label{fig:problem}
\end{figure} 

However, since IoT devices are resource-constrained (e.g., limited storage and computing resources), it is difficult to provide sufficient resources for training resource-intensive ViT models.
Therefore, it has become mainstream to assist model training with the help of cloud/edge servers~\cite{Ma@Mobility,Ding@Resource,Wang@a,Tuli@Dynamic,Chen@EnergyEfficient,Wang@Fast}.
In general, the cloud server first trains the ViT model and then sends it to the IoT device for deployment and updating, as illustrated in Fig.~\ref{fig:problem}.
The ViT model contains a large number of parameters (e.g., ViT~\cite{Carion@End} contains 41 million parameters), which leads to transmitting a large number of model parameters when the cloud server assists in deploying and updating the model.
In particular, in the construction of the smart city, when a cloud server assists thousands of IoT devices, the transmission of a large number of model parameters will lead to heavy network load and difficulty in online deployment and updating of ViT models.
In addition, existing ViT models contain a large number of floating-point operations (FLOPs), making them difficult to deploy on resource-constrained IoT devices.
Therefore, to facilitate the cloud server to assist IoT devices in deploying and updating the model, it is necessary to reduce the number of model parameters and FLOPs, especially in today's Internet of Everything era.
To this end, this paper aims to implement a transmission-friendly ViT model for IoT devices to reduce the number of model parameters and FLOPs while improving model performance.

First, this paper proposes a hybrid layer consisting of nonlearnable (NL) modules and a pointwise convolution to replace the multihead attention (MHA) of the standard ViT.
The NL module consists of maximum (max) pooling and average (avg) pooling.
Additionally, the NL module can easily and parallelly extract multitype and multiscale features to improve the transformer performance, e.g., using $3\times 3$, $5\times 5$, or $7\times 7$ max/avg-pooling. 
The pointwise convolution consists of $1\times 1$ convolutions with only a few learnable parameters.
Therefore, replacing the MHA with the proposed hybrid layer can greatly reduce the number of model parameters and FLOPs.

Then, this paper introduces the group convolution for channel sparse connection to reduce the number of parameters in the feedforward network of the ViT.
Group convolution can significantly reduce the number of parameters by ensuring that each convolution operates only on the corresponding group of input channels.
Motivated by~\cite{Zhang@ShuffleNet}, we introduce the channel shuffle operation to allow group convolution to obtain input data from different groups; that is, the input and output channels can be fully related.
We also present a tradeoff between the number of model parameters and model performance analyzed from the experimental results.

By introducing the hybrid layer and group convolution into the existing ViT, we propose a {\underline{T}}ransmission-{\underline{F}}riendly vision Transf{\underline{ormer}} (TFormer), which has fewer parameters and FLOPs while also achieving higher performance.
It can be deployed on a large scale on IoT devices with cloud-assisted training.
Since our \ourmethod{} is similar in structure to the existing ViT, we call \ourmethod{} a kind of ViT.

Finally, this paper conducts extensive experiments on the ImageNet-1K, MS COCO, and ADE20K datasets, and the experimental results show that our proposed TFormer achieves strong performance on the recognition tasks of image classification, object detection, and semantic segmentation. 
For example, when the number of model parameters is similar, our proposed TFormer achieves 41.2 average precision (AP) on the MS COCO dataset, which surpasses the previous state-of-the-art result by +2.3 AP.
ADE20K semantic segmentation obtains 41.8 mean intersection over union (mIoU), an improvement of +2.3 mIoU over the previous state-of-the-art results.
In addition, we also demonstrate the advantages of \ourmethod{} in the number of model parameters and FLOPs, making it more suitable for deployment on resource-constrained IoT devices. 

\noindent In summary, our main contributions are as follows:

\vspace{2pt}
\noindent\textbf{--} This paper presents a new pathway towards efficient ViT models for IoT devices, comprising a novel hybrid layer and a novel feedforward network with group-wise connections. 

\vspace{2pt}
\noindent\textbf{--} The key component of the proposed hybrid layer is the nonlearnable module with multiple non-parametric operations (i.e., pooling operators with various kernel sizes) in parallel to extract multitype and multiscale features, while at the same time reducing the model parameters and FLOPs substantially. 

\vspace{2pt}
\noindent\textbf{--} Extensive experiments conducted on the ImageNet, MS COCO, and ADE20K datasets verify the effectiveness of the proposed method over classification, object detection, and semantic segmentation tasks.

The remainder of the paper is organized as follows. Section~\ref{ref:2-related} reviews the related work on cloud-assisted approaches and ViT models. Section~\ref{ref-proposedapproach} describes the proposed TFormer model in detail. Section~\ref{ref-experiments} presents our evaluation results, and Section~\ref{ref-conclusion} concludes the paper.

\section{Related Work}\label{ref:2-related} 
In this section, we introduce the cloud-assisted approach and the ViT models that are most relevant to this paper.

\subsection{Cloud-assisted Approach}
Due to the limited resources of IoT devices, it is difficult to provide sufficient computing and storage resources for training high-performance deep neural network models (e.g., deep convolutional neural networks and transformers).
To this end, it has become the mainstream to first use well-resourced cloud/edge servers to assist in training high-performance models~\cite{Xu@Energy,Ding@Cognitive,Teerapittayanon@Distributed,Han@Accelerating,Blakeney@Parallel,Ding@Towards,HiTDL,8943178,AppATP}.
For example, inspired by~\cite{Yosinski@How}, Ding et al.~\cite{Ding@Cognitive, Ding@A} proposed training a complex neural network model on a cloud server and a simple neural network model on an edge server and improving the performance of the latter by sharing some layer parameters of the complex neural network model with the simple neural network model.
Teerapittayanon et al.~\cite{Teerapittayanon@Distributed} proposed a distributed deep neural network architecture consisting of cloud servers, edge servers and IoT devices.
Similarly, Kang et al.~\cite{Kang@Neurosurgeon} proposed dividing the neural network model into two parts, which are run on cloud servers and IoT devices.

Many studies have also made innovative contributions to adapting IoT devices.
For example, to adapt to the dynamically changing available resources of IoT devices, Han et al.~\cite{Han@An} proposed deploying multiple model variants on IoT devices.
Fang et al.~\cite{Fang@NestDNN} proposed making multiple model variants share parameters to save the limited storage resources of the IoT device.
Additionally, to reduce the number of model parameters when the cloud server assists in training the IoT neural network model, many researchers proposed using model compression~\cite{Han@Deep} or knowledge distillation~\cite{Hinton@Distilling} to reduce the amount of model parameter transmission.
For example, Laskaridis et al.~\cite{Stefanos@SPINN} proposed using a model compression technique to reduce the number of model parameters during the interaction between the cloud server and the IoT device.

Different from the above methods, our method focuses on reducing the number of model parameters and floating-point operations by analyzing the components of the model in detail and introduces pooling techniques and group convolutions to achieve this goal.
Model compression techniques can further reduce the number of model parameters based on our method.
The study of adapting IoT devices~\cite{Han@An,Fang@NestDNN} can also be built on our method.

\begin {figure*}[t]
\centering
\includegraphics[width=0.9\linewidth]{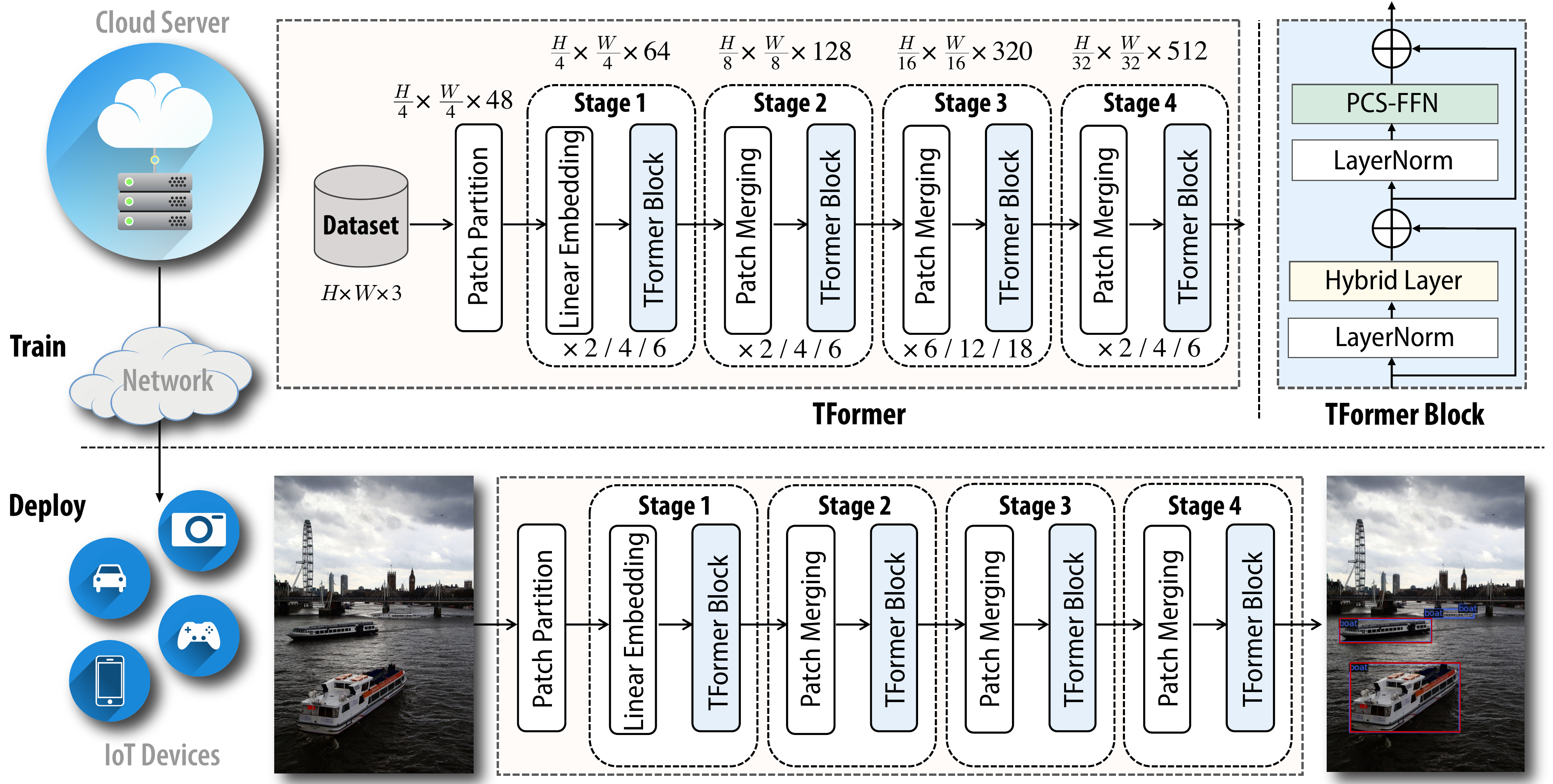}
\caption {Overview of the proposed framework.} 
\label{fig:TFormer_framework}
\end{figure*} 

\subsection{Vision Transformers}
Considering the great success of the transformer~\cite{Vaswani@Attention} in the natural language processing field, the application of the transformer architecture to the vision field has attracted the attention of a large number of scholars and achieved attractive results~\cite{Dosovitskiy@An,Liu@Swin,Touvron@Training,Wu@CvT,Yuan@Incorporating,Yuan@Tokens}.
For example, Dosovitskiy et al.~\cite{Dosovitskiy@An} pioneered the standard transformer to process images directly.
They split the image into patches and provide a sequence of linear embeddings of these patches as input to the transformer.
Liu et al.~\cite{Liu@Swin} proposed a hierarchical transformer (swin transformer) to support large-scale variations of visual entities and high-resolution pixels in images.
To improve the convergence rate of the transformer, Touvron et al.~\cite{Touvron@Training} proposed a teacher-student strategy for the transformer. 
Wu et al.~\cite{Wu@CvT} introduced convolution in the standard transformer to improve the performance of the transformer so that the transformer containing the convolution affords both the advantages of the convolution and the standard transformer.
Similarly, Yuan et al.~\cite{Yuan@Incorporating} also introduced convolution in the standard transformer to improve the transformer's performance in vision tasks. 

Accordingly, existing vision transformers achieve high performance by leveraging the multihead attention module to capture the global information of data.
However, due to the quadratic complexity of the multihead attention module, it is difficult for existing ViT models to be widely deployed in resource-constrained IoT devices.
In addition, the feedforward network module included in the existing ViT model contains a large number of parameters, which makes it necessary to transmit a large number of model parameters during cloud-assisted deployment and updating.
This will hinder the wide deployment of the ViT model in IoT devices in the era of the Internet of Everything,  where bandwidth resources are tight.
To this end, we analyze the components of the ViT model in detail from the perspective of cloud-assisted deployment to reduce the number of model parameters and floating-point operations, as well as maintain the performance of the model.

\vspace{3pt}
Existing efforts toward improving the efficiency of ViT models can be broadly classified into three categories.
The first group of methods focuses on reducing the complexity of the attention module by imposing the locality of input images adaptively \cite{wang2021not,meng2022adavit}.
The second group applies pruning methods to remove unimportant components (e.g., partial channels) \cite{yu2022width} or inputs (i.e., patches) \cite{tang2022patch} to a ViT model.
The third group of methods uses neural architecture search (NAS) techniques to design efficient ViT models by optimizing architectural hyperparameters, such as channels and depth \cite{chen2021autoformer,chen2021glit}. 
On one hand, the improvements in model efficiency provided by adaptive attention mechanisms are still insufficient for tasks with high-resolution imagery (e.g., object detection, segmentation, etc.).
On the other hand, despite the promising results, both pruning and NAS-based approaches are computationally expensive, requiring days to weeks on a cluster of GPUs to execute the methods.

\section{Design of the Proposed Approach} \label{ref-proposedapproach}
In this section, we first introduce the overview of the proposed framework and motivation.
We then provide the detailed design of our proposed hybrid layer and partially connected and shuffled feedforward network (PCS-FFN).

\subsection{Overview \label{sec:overview}}
\begin{table}[ht]
\centering
\caption{Specific design details of three TFormer variants.\label{tab:model_schema}}
\resizebox{.42\textwidth}{!}{%
\begin{tabular}{@{\hspace{2mm}}cccccc@{\hspace{2mm}}}
\toprule
\multirow{2}{*}{Stage} & \multirow{2}{*}{\#Tokens} & \multirow{2}{*}{Specifications} & \multicolumn{3}{c}{TFormer} \\ \cmidrule(l){4-6} 
 &  &  & \multicolumn{1}{c|}{S} & \multicolumn{1}{c|}{M} & L \\ \midrule
\multirow{5}{*}{1} &  \multirow{5}{*}{$\frac{\mbox{H}}{4}\times\frac{\mbox{W}}{4}$} & Patch size & \multicolumn{3}{c}{7$\times$7, stride 4} \\ \cmidrule(l){3-6} 
 &  & Embed. dim. & \multicolumn{3}{c}{64} \\ \cmidrule(l){3-6} 
 &  & FFN & \multicolumn{3}{c}{ratio 4, groups 2} \\ \cmidrule(l){3-6} 
 &  & \#Layers & \multicolumn{1}{c|}{2} & \multicolumn{1}{c|}{4} & 6 \\ \midrule
\multirow{5}{*}{2} &  \multirow{5}{*}{$\frac{\mbox{H}}{8}\times\frac{\mbox{W}}{8}$} & Patch size & \multicolumn{3}{c}{3$\times$3, stride 2} \\ \cmidrule(l){3-6} 
 &  & Embed. dim. & \multicolumn{3}{c}{128} \\ \cmidrule(l){3-6} 
 &  & FFN & \multicolumn{3}{c}{ratio 4, groups 2} \\ \cmidrule(l){3-6} 
 &  & \#Layers & \multicolumn{1}{c|}{2} & \multicolumn{1}{c|}{4} & 6 \\ \midrule
\multirow{5}{*}{3} &  \multirow{5}{*}{$\frac{\mbox{H}}{16}\times\frac{\mbox{W}}{16}$} & Patch size & \multicolumn{3}{c}{3$\times$3, stride 2} \\ \cmidrule(l){3-6} 
 &  & Embed. dim. & \multicolumn{3}{c}{320} \\ \cmidrule(l){3-6} 
 &  & FFN & \multicolumn{3}{c}{ratio 4, groups 2} \\ \cmidrule(l){3-6} 
 &  & \#Layers & \multicolumn{1}{c|}{6} & \multicolumn{1}{c|}{12} & 18 \\ \midrule
\multirow{5}{*}{4} &  \multirow{5}{*}{$\frac{\mbox{H}}{32}\times\frac{\mbox{W}}{32}$} & Patch size & \multicolumn{3}{c}{3$\times$3, stride 2} \\ \cmidrule(l){3-6} 
 &  & Embed. dim. & \multicolumn{3}{c}{512} \\ \cmidrule(l){3-6} 
 &  & FFN & \multicolumn{3}{c}{ratio 4, groups 2} \\ \cmidrule(l){3-6} 
 &  & \#Layers & \multicolumn{1}{c|}{2} & \multicolumn{1}{c|}{4} & 6 \\ \midrule
\multicolumn{3}{c}{Parameters (M)} & \multicolumn{1}{c|}{8} & \multicolumn{1}{c|}{14} & 20 \\ \midrule
\multicolumn{3}{c}{Multi-Adds (G)} & \multicolumn{1}{c|}{1.2} & \multicolumn{1}{c|}{2.2} & 3.2 \\ \bottomrule
\end{tabular}%
}
\end{table}

The overall framework first trains a vision transformer (ViT) model on the cloud server.
Then, the cloud server sends the trained model to IoT devices to provide a variety of convenient services, such as object detection and segmentation services.
See Fig.~\ref{fig:TFormer_framework} for a pictorial overview. 

The goal of this paper is to develop a transmission-friendly ViT model while ensuring its performance.
To fulfill this goal, we first propose a compact yet effective ViT model, dubbed \ourmethod{}, tailored for IoT applications. 
As depicted in Fig.~\ref{fig:TFormer_framework} (top-right), the main computational block of \ourmethod{} comprises a \emph{hybrid} layer and a \emph{partially connected and shuffled} feedforward network (PCS-FFN), replacing the multihead attention (MHA) layer and the standard FFN in existing ViTs, respectively. 
The design principles of the hybrid layer and PCS-FFN are carefully explained and empirically validated in Section~\ref{ref:multiscalenonlearnable} and Section~\ref{ref:partialFFN}, respectively.  
Subsequently, we develop three variants of \ourmethod{} with an increasing model capacity, i.e., \ourmethod{}-S, \ourmethod{}-M, and \ourmethod{}-L.
The specific configuration details of the three variants are provided in Table~\ref{tab:model_schema}.

\subsection{Motivation}
Generally, there are two aspects to measure whether a model is suitable for running on resource-constrained IoT devices, namely, the number of model parameters and the number of floating-point operations (FLOPs) required to run the model. To be consistent with prior work, we consider the number of floating-point operations for multiplication and addition together as one FLOP \cite{He@Deep}. 

On the one hand, the number of model parameters affects the deployment of the model in two ways: (i) the quantity of data required to be transmitted between cloud and IoT devices and (ii) the size of the model.
First, we aim to train a ViT on the cloud server, which is then sent to IoT devices for deployment.
In the Internet of Everything era, considering the online deployment and updating of models on hundreds of millions of IoT devices, it is necessary to reduce the quantity of data sent by the cloud server to IoT devices.
Second, the number of model parameters determines both the storage space and memory resources of the IoT device that will be occupied by the model.

On the other hand, the number of FLOPs signifies the computational complexity of a ViT model, where a higher value of FLOPs typically results in higher power consumption and longer inference latency \cite{li2021hw,benmeziane2021comprehensive}. Particularly, IoT devices are mostly constrained by resources, such that the level of computational complexity (FLOPs) of a ViT model is especially important. 

Therefore, to facilitate cloud-assisted deployment and updating of models on IoT devices, it is necessary to reduce the number of parameters and the number of FLOPs of the ViT model \emph{simultaneously}.

\subsection{Design of Hybrid Layer} \label{ref:multiscalenonlearnable}
\noindent \textbf{Preliminaries.} The multihead attention (MHA) is one of the cornerstones in existing ViT models \cite{Vaswani@Attention,Yuan@Tokens,Liu@Swin,Dosovitskiy@An,Park@How}.
Undoubtedly, MHA is important for model performance owing to its ability to extract nonlocal dependencies. Nevertheless, the steep computational overhead has greatly restricted its application to resource-constrained hardware, such as IoT devices. 

Specifically, let us consider an input ${\bf{X}} \in \mathds{R}^{N\times D}$, where $N$ is the size of the input (i.e., number of pixels) and $D$ is the embedding dimension (i.e., number of channels).
Assuming the number of attention heads is $h$, then for each head $i$, the input $\bf{X}$ is embedded into a \emph{query} (${\bf{Q}}_i$), \emph{key} (${\bf{K}}_i$), and \emph{value} (${\bf{V}}_i$), as follows:
\begin{equation}
\begin{array}{l}
{\bf{Q}}_i = {\bf{X}}{\bf{W}}_i^Q, \hspace{1em}{\bf{K}}_i = {\bf{X}}{\bf{W}}_i^K, \hspace{1em}{\bf{V}}_i = {\bf{X}}{\bf{W}}_i^V,
\end{array}
\label{eq:QKV}
\end{equation}

\noindent where ${\bf{W}}_i^Q$, ${\bf{W}}_i^K$, and ${\bf{W}}_i^V \in \mathds{R}^{D\times D/h}$ are learnable parameters for $i={\emph{1}},\ldots, h$. The attention for the $i$th head is then computed as follows:
\begin{equation}
\begin{array}{l}
\text{Attention} ({\bf{Q}}_i, {\bf{K}}_i, {\bf{V}}_i) = softmax\Big(\frac{{\bf{Q}}_i{\bf{K}}_i^\intercal}{\sqrt{\frac{D}{h}}}\Big){\bf{V}}_i
\end{array}.
\label{eq:Attention}
\end{equation}

\noindent Then, the outputs from all heads are concatenated along the channel dimension:
\begin{equation}
\begin{array}{l}
\begin{split}
\text{[head$_1$, $\ldots$, head$_h$]} = \text{Concat}\big(&\text{Attention} ({\bf{Q}}_1, {\bf{K}}_1, {\bf{V}}_1), \ldots,\\ & \text{Attention} ({\bf{Q}}_h, {\bf{K}}_h, {\bf{V}}_h)\big)\\
\end{split}
\end{array}
\label{eq:concat_attn}
\end{equation}

\noindent Finally, the concatenated outputs are projected via another learnable parameter ${\bf{W}} \in \mathds{R}^{D\times D}$, yielding the final outputs of the MHA as follows:
\begin{equation}
\begin{array}{l}
\text{MHA} ({\bf{X}}) = \text{[head$_1$, $\ldots$, head$_h$]}{\bf{W}}
\end{array}
\label{eq:MultiAttention}
\end{equation}

\noindent Accordingly, the number of parameters of MHA can be derived as:

\begin{equation}
\begin{array}{l}
P_{\mbox{\tiny MHA}} = \underbrace{3\sum_{i=1}^{h}D\times\frac{D}{h}}_{{\bf{W}}^Q, {\bf{W}}^K, {\bf{W}}^V} + \underbrace{D\times{D}}_{\bf{W}} = 4D^{2},
\end{array}
\label{eq:mha_params}
\end{equation}

\noindent which is \emph{quadratic} to the embedding dimension $D$. Considering typical embedding dimension values (i.e., $D \in \{128, 320, 512\}$) used in existing ViTs \cite{Dosovitskiy@An,Touvron@Training}, the resulting number of parameters can be prohibitive to transmit between cloud and IoT devices. Similarly, we can derive the number of FLOPs of MHA as follows:

\begin{equation}
\begin{array}{l}
F_{\mbox{\tiny MHA}} = \underbrace{3ND\frac{D}{d}}_{\small\mbox{Eq~\ref{eq:QKV}}} + \underbrace{N\frac{D}{h}N + NN\frac{D}{h}}_{\small\mbox{Eq~\ref{eq:Attention}-\ref{eq:concat_attn}}} + \underbrace{NDD}_{\small\mbox{Eq~\ref{eq:MultiAttention}}}.
\end{array}
\label{eq:mha_flops}
\end{equation}

\noindent Since the input size is typically much greater than the embedding dimension (i.e., $N \gg D$), $F_{\mbox{\tiny MHA}}$ is approximately \emph{quadratic} to the input size $N$. Given the input sizes of the considered tasks (e.g., 1200$\times$800 for object detection), one computation of MHA can already be too expensive for IoT devices. 

\begin{figure}[t]
    \centering
    \includegraphics[width=.95\linewidth]{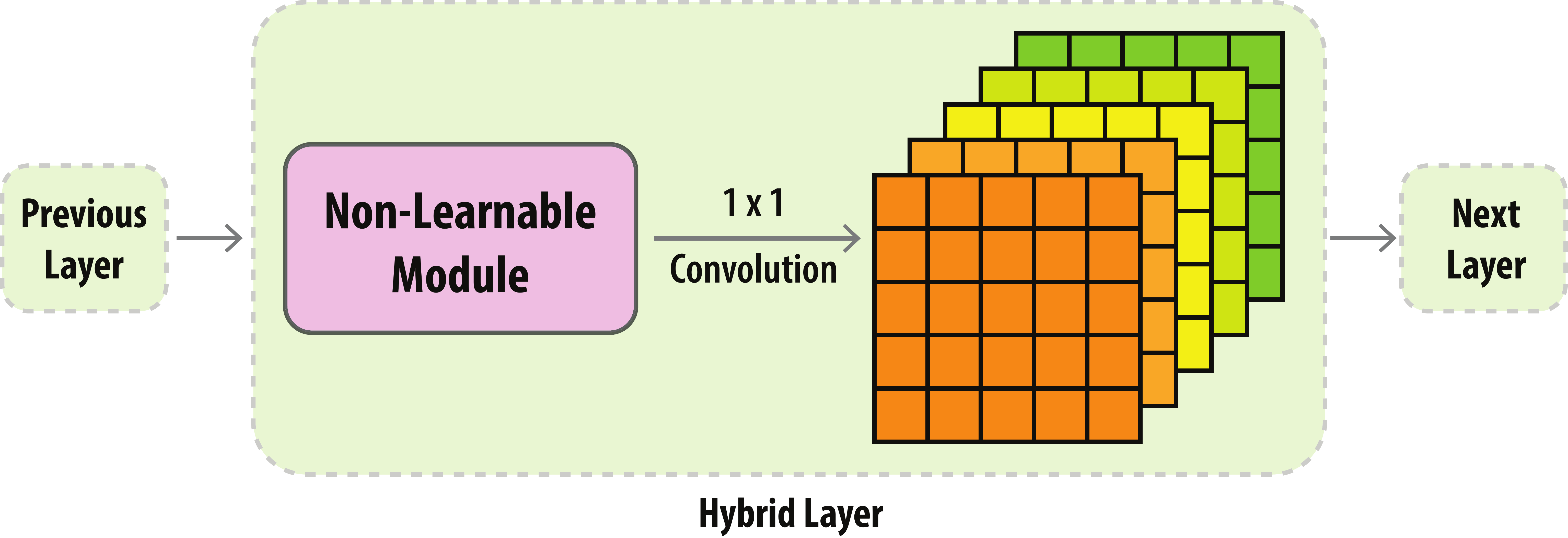}
    \caption{{\bf{Hybrid layer:}} It first uses the nonlearnable module to extract multitype and multiscale features and then uses pointwise ($1\times 1$) convolution to enable these features to establish information interactions between channels.}
    \label{fig:hybrid_layer}
\end{figure}

\vspace{2pt}
\noindent \textbf{Design principles.} To this end, we aim to propose a novel layer to replace MHA for reducing both the number of model parameters and FLOPs while maintaining similar model performance.
As explained by the original authors~\cite{Vaswani@Attention}, the role of MHA is to enable ViT models to jointly focus on information from different representation subspaces at different locations.
From this point of view, it can be equivalent to using multiple types of filters to extract diverse features.
Under this assumption, we propose a hybrid layer that contains only a few learnable parameters but can extract multitype and multiscale features.
As depicted in Fig.~\ref{fig:hybrid_layer}, the proposed hybrid layer comprises a nonlearnable (NL) module and a pointwise (i.e., 1$\times$ 1) convolution.
In this hybrid layer, the NL module is used to extract diverse spatial information, while the pointwise convolution is learned to recombine the multitype and multiscale information.

\begin {figure}[t]
\centering
\begin{subfigure}[b]{0.45\textwidth}
\centering
\includegraphics[width=.95\linewidth]{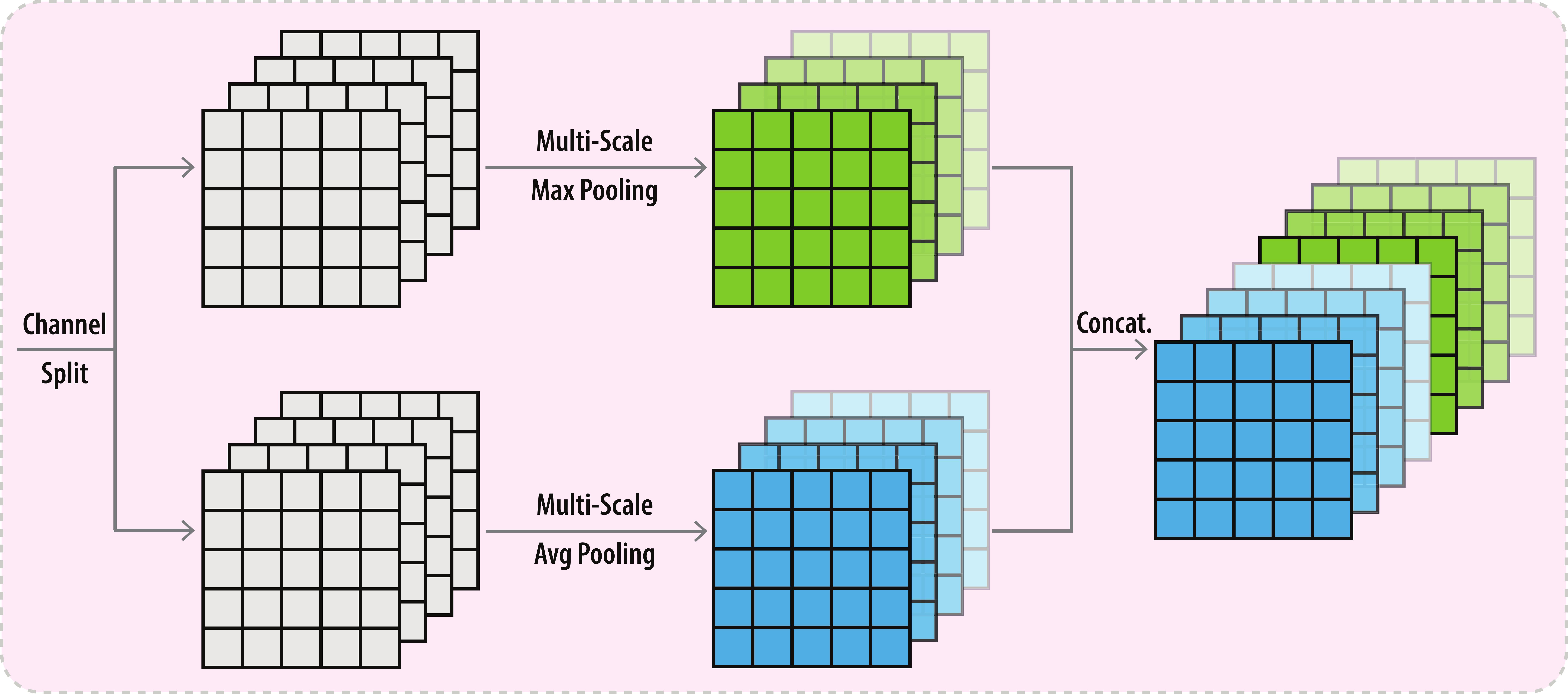}
\caption{Nonlearnable modules\label{fig:nl_modules}}
\end{subfigure} \\ \vspace{3pt}
\begin{subfigure}[b]{0.45\textwidth}
\centering
\includegraphics[width=.95\linewidth]{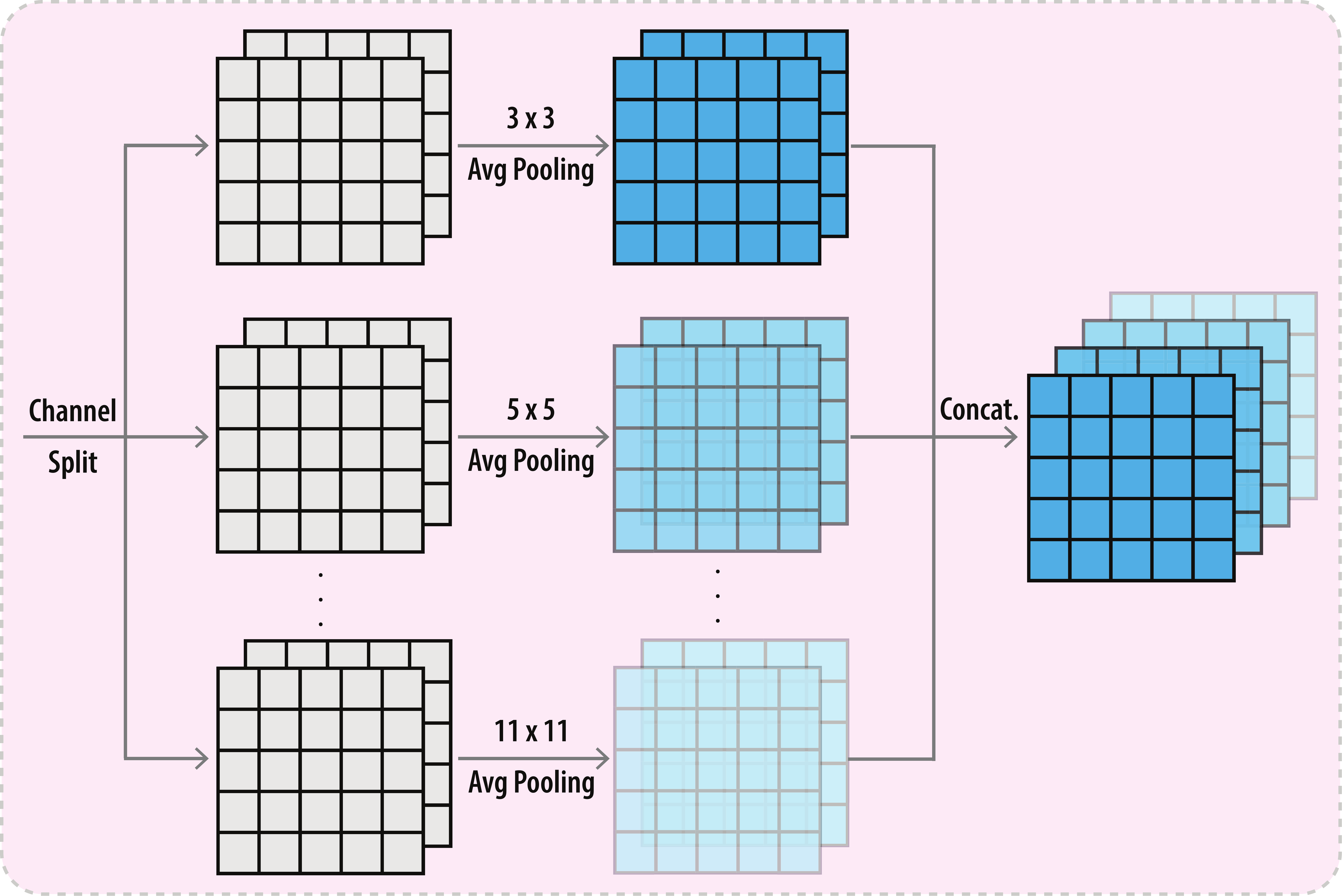}
\caption{Multiscale average pooling\label{fig:avg_ms}}
\end{subfigure}
\caption {(a) Our nonlearnable modules comprise nonparametric pooling operations, i.e., avg and max poolings. (b) We allow each pooling operation to have multiple kernel size settings in parallel to capture multiscale features.}
\label{fig:non_learnable_modules}
\end{figure} 

As depicted in Fig.~\ref{fig:nl_modules}, our NL module comprises a max-pooling operation and an avg-pooling operation.
The max-pooling operation calculates the maximum value of feature patches, which is known to be invariant to a small amount of translation.
In parallel, the avg-pooling operation calculates the mean value for feature patches. 
Furthermore, we allow each pooling operation to have multiple kernel size settings in parallel to capture multiscale features, as depicted in Fig.~\ref{fig:avg_ms}. 
We set the padding accordingly to maintain the same spatial size of inputs. 
More specifically, let's consider that there are one input and one output feature map. Then the size of the extracted output feature map is calculated as $N = [(W-K+2P)/S]+1 $, where $W$ is the size of the input feature map; $K$ is the pooling size (e.g., $3\times3$, $5\times5$, etc.), $P$ is the padding size; $S$ is the stride. To maintain the same sizes between input $W$ and output $N$ feature maps under different pooling size $K$, we first set the stride $S$ to one then adjust the padding $P$ accordingly. For instance, we set $P$ to 1 and 2 for pooling sizes of $3\times3$ and $5\times5$ respectively.
Technically, one may have as many NL operations in parallel as one wishes, and we reserve this part of the work as in our future studies.

Note that we split the inputs along the channel dimension for each NL operator and for each kernel size within that operator to further reduce the computations. 
Specifically, assuming that the number of input channels is $D$ and the number of considered kernel sizes is $m$, such a channel splitting operation will maintain the same number of output channels (as inputs) of $D$, instead of $2*D*m$ for the case without channel splitting. 
Then, the number of learnable parameters of the proposed hybrid layer is simply the number of parameters of the pointwise convolution, i.e., $P_{\mbox{\tiny NL Layer}} = D^2$ assuming an equal number of input and output channels. 
Hence, the ratio of the number of parameters between the MHA and our hybrid layer is as follows:
\begin{equation}
\begin{array}{l}
R_{\mbox{\tiny P}} = \frac{P_{\mbox{\tiny MHA}}}{P_{\mbox{\tiny NL Layer}}} = \frac{4D^2}{D^2} = 4.
\end{array}
\label{eq:parameter_ratio}
\end{equation}

\noindent Apparently, compared to MHA, our hybrid layer leads to a 4$\times$ reduction in the number of parameters under the assumption of equal embedding dimensions and channels. 

The number of FLOPs of the proposed hybrid layer can be computed as:
\begin{equation}
\begin{array}{l}
F_{\mbox{\tiny NL Layer}} = \underbrace{DNk^2}_{\mbox{\small NL modules}} + \underbrace{D^2N}_{\mbox{\small 1}\times \mbox{\small 1}~\mbox{\small Conv.}},
\end{array}
\label{eq:nl_layer_flops}
\end{equation}

\noindent where $k$ is the kernel size, $N$ is the input size (i.e., number of pixels), and $D$ is the number of input/output channels. Since the kernel size is typically much smaller than the number of channels (i.e., $k \ll D$), $F_{\mbox{\tiny NL Layer}}$ is approximately quadratic to the number of channels $D$. Thus, the ratio of the number of FLOPs between MHA and our hybrid layer can be approximated as follows: 
\begin{equation}
\begin{array}{l}
R_{\mbox{\tiny F}} = \frac{F_{\mbox{\tiny MHA}}}{F_{\mbox{\tiny NL Layer}}} \sim \frac{N^2}{D^2}.
\end{array}
\label{eq:flops_ratio}
\end{equation}

\noindent Given the vision tasks considered in this work (e.g., object detection, segmentation, etc.), the input size (e.g., $N = 1200\times800$ for object detection) is much greater than the number of channels (e.g., $D = 512$), and the resulting FLOPs ratio is expected to be much greater than 1 (i.e., $R_{\mbox{\tiny F}} \gg 1$). Hence, our hybrid layer also leads to significant savings in the number of FLOPs compared to MHA. 

\begin{table}[ht]
\centering
\caption{Ablation studies of our nonlearnable modules. All variants have similar number of parameters and FLOPs. \label{tab:abl_nonlearnable}}
\resizebox{.48\textwidth}{!}{%
\begin{tabular}{@{\hspace{2mm}}cc|ccccc|c@{\hspace{2mm}}}
\toprule
\multicolumn{2}{c|}{Non-learnable   operator} & \multicolumn{5}{c|}{Scale} & \multirow{2}{*}{\begin{tabular}[c]{@{}c@{}}ImageNet\\Top-1 Acc\end{tabular}} \\ \cmidrule(l){1-2} \cmidrule(l){3-7}
Avg. Pool & Max Pool & 3$\times$3 & 5$\times$5 & 7$\times$7 & 9$\times$9 & 11$\times$11 &  \\ \midrule
\checkmark &  & \checkmark &  &  &  &  & 77.2\% \\ 
\checkmark &  &  & \checkmark &  &  &  & 77.2\% \\ 
\checkmark &  &  &  & \checkmark &  &  & 77.1\% \\ 
\checkmark &  &  &  &  & \checkmark &  & 76.8\% \\ \midrule
\checkmark &  & \checkmark & \checkmark &  &  &  & 77.2\% \\
\checkmark &  & \checkmark & \checkmark & \checkmark &  &  & 77.3\% \\ 
\checkmark &  & \checkmark & \checkmark & \checkmark & \checkmark &  & 77.5\% \\ 
\checkmark &  & \checkmark & \checkmark & \checkmark & \checkmark & \checkmark & 77.7\% \\ \midrule
\checkmark & \checkmark & \checkmark &  &  &  &  & 77.6\% \\ 
\checkmark & \checkmark & \checkmark & \checkmark & \checkmark & \checkmark & \checkmark & 78.0\% \\ \bottomrule
\end{tabular}%
}
\end{table}

\begin{figure*}[t]
    \centering
    \begin{subfigure}[b]{0.32\textwidth}
    \centering
    \includegraphics[width=.95\textwidth]{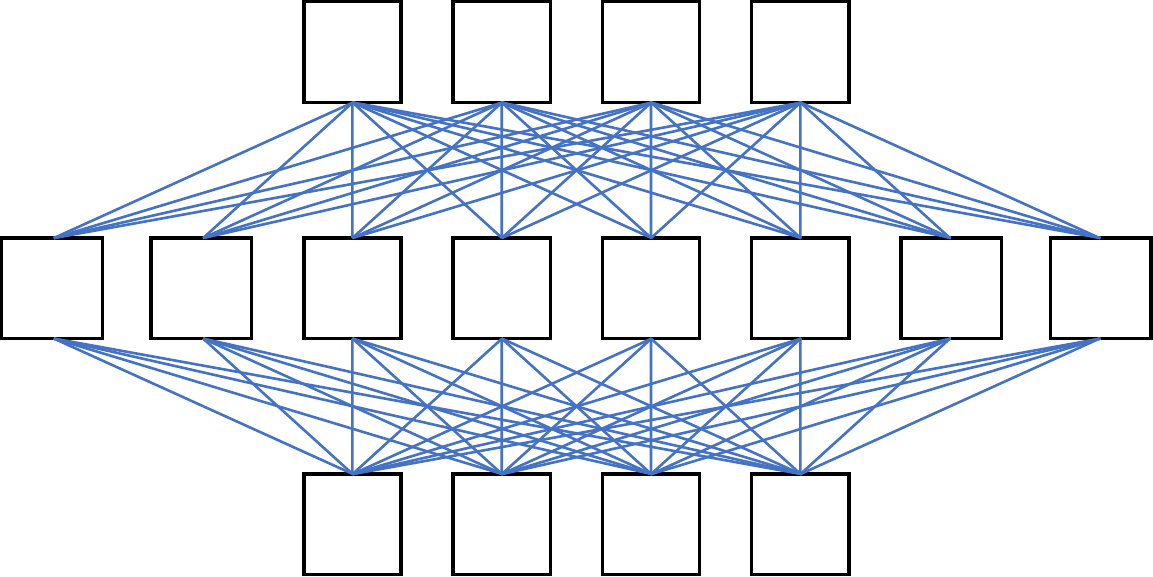}
    \caption{Standard FFN ($r$=2) \label{fig:standardFFN}}
    \end{subfigure}\hfill
    \centering
    \begin{subfigure}[b]{0.32\textwidth}
    \centering
    \includegraphics[width=.95\textwidth]{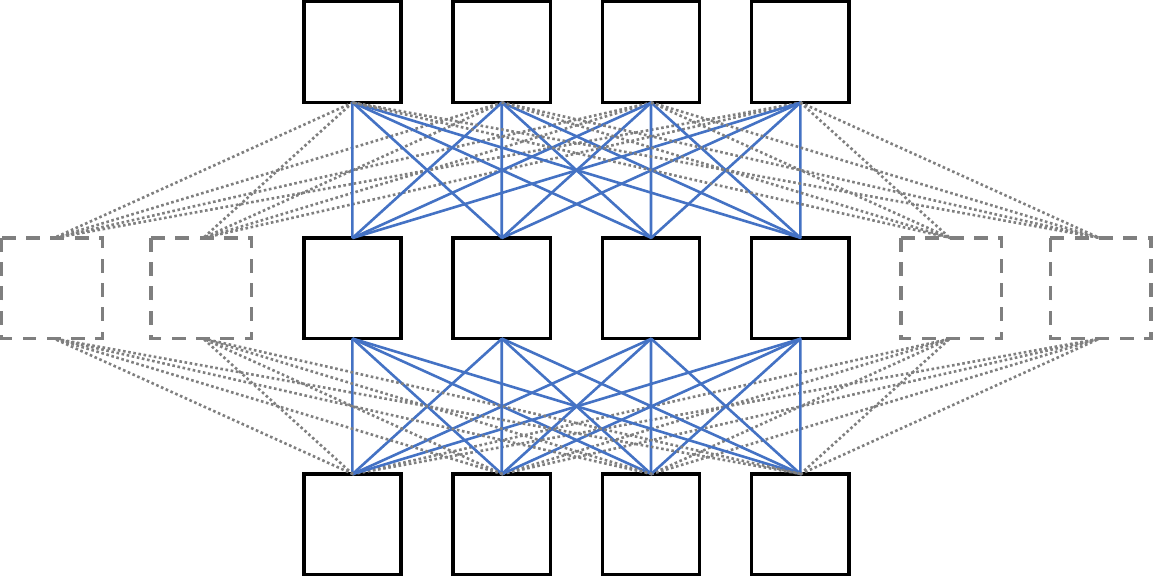}
    \caption{Standard FFN ($r$=1) \label{fig:FFNwithratio}}
    \end{subfigure}\hfill
    \centering
    \begin{subfigure}[b]{0.32\textwidth}
    \centering
    \includegraphics[width=.8\textwidth]{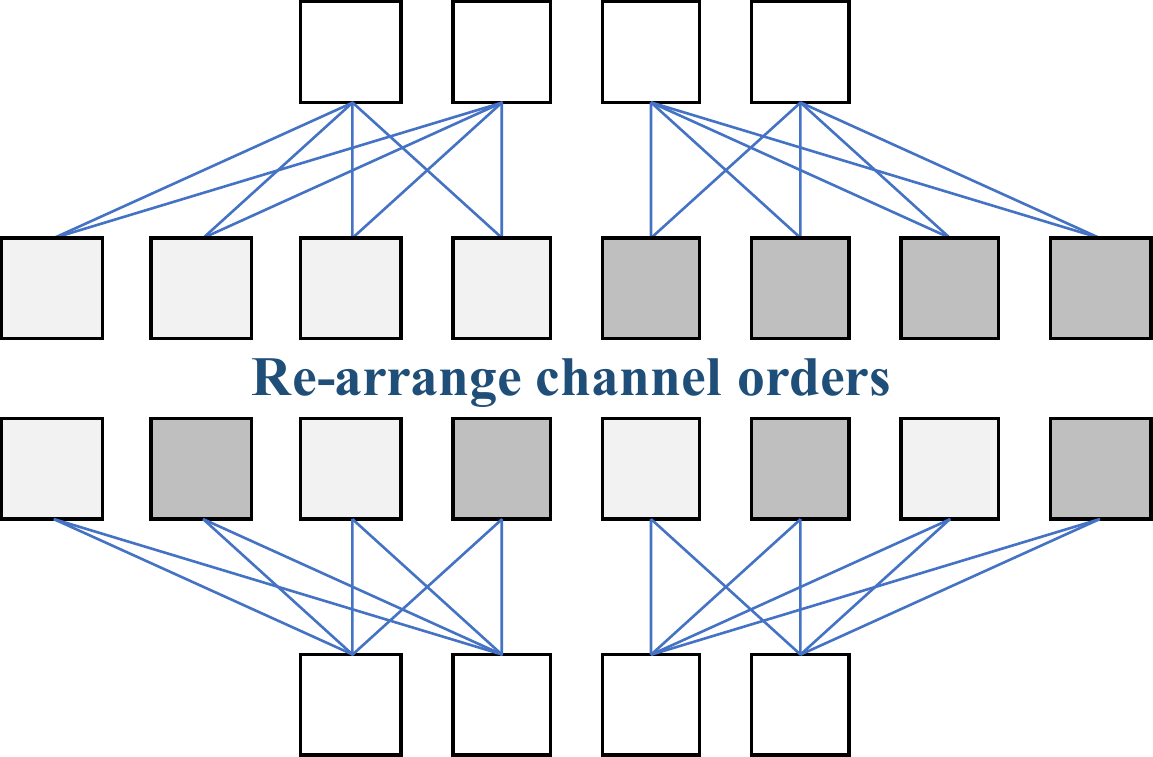}
    \caption{Our PCS-FFN ($r$=2, g=2) \label{fig:proposedPCS-FFN}}
    \end{subfigure}
    \caption{(a) Standard FFN comprises two fully connected linear layers to first expand the number of channels by a ratio $r$ and then compress them back to the original number of channels. (b) One straightforward way to reduce model parameters is by reducing the expansion ratio $r$. (c) Instead, we propose using partially connected linear layers with rearranging channels between the two layers. Note that both (b) and (c) achieve 2$\times$ savings in parameters.\label{fig:result_overview}}
\end{figure*}

\vspace{2pt}
\noindent \textbf{Effectiveness.} Next, we provide empirical validations of our hybrid layer. Specifically, we consider the image classification task on the ImageNet-1K dataset \cite{Russakovsky@ImageNet}; we replace the MHA layer with our hybrid layer in ViTs \cite{Dosovitskiy@An}; and compare different variants of our hybrid layer. 
Experimental results are summarized in Table~\ref{tab:abl_nonlearnable}.
We observe that both adding more types of operations and adding more scales (i.e., kernel sizes) are beneficial to the model performance without additional learnable parameters. 
These results provide the empirical basis for the design of our hybrid layer. 

\subsection{Design of PCS-FFN} \label{ref:partialFFN}
\noindent \textbf{Preliminaries.} In addition to MHA, the other main computational bottleneck in existing ViT models is the feedforward network (FFN), which is usually placed immediately after the MHA layer (see Fig.~\ref{fig:TFormer_framework} top-right). 
An FFN comprises two fully connected (linear) layers with an activation in between, as shown in Fig.~\ref{fig:standardFFN}.
For an input ${\bf{X}_{\mbox{\tiny in}}}\in \mathds{R}^{N\times D}$, the first linear layer expands the number of channels to a ratio of $r$, yielding an intermediate output ${\bf{X}_{\mbox{\tiny inter}}}\in \mathds{R}^{N\times rD}$. 
Then, the second linear layer compresses channels back to their original dimensions, outputting ${\bf{X}_{\mbox{\tiny out}}}\in \mathds{R}^{N\times D}$.
For a standard FFN, the number of parameters is $P_{\mbox{\tiny FFN}}=D\times rD + rD\times D=2rD^2$, and the number of FLOPs is $F_{\mbox{\tiny FFN}}=2rND^2$.
In existing ViT models, the expansion ratio $r$ is typically set to 4.

\begin{table}[ht]
\centering
\caption{Ablative study on the proposed FFN. Our FFN is highlighted in bold. All results are on ImageNet-1K. \label{tab:abl_ffn}}
\resizebox{.45\textwidth}{!}{%
\begin{tabular}{@{\hspace{2mm}}lccc@{\hspace{2mm}}}
\toprule
Method & \#Params & \#MAdds & Top-1 Acc \\ \midrule
Standard FFN ($r$=1) \cite{Dosovitskiy@An} & 6.0M & 0.83G & 70.8\% \\
Ghost FFN ($r$=4) \cite{Han@GhostNet} & 6.2M & 0.87G & 71.8\% \\
\textbf{PCS-FFN ($r$=4, g=4)} & \textbf{6.0M} & \textbf{0.83G} & \textbf{73.2\%} \\ 
\midrule
Standard FFN ($r$=2) \cite{Dosovitskiy@An} & 8.4M & 1.2G & 75.0\% \\
Ghost FFN ($r$=2) \cite{Han@GhostNet} & 8.5M & 1.3G & 75.5\% \\
\textbf{PCS-FFN ($r$=4, g=2)} & \textbf{8.4M} & \textbf{1.2G} & \textbf{76.1\%} \\ 
\midrule
Standard FFN ($r$=4) & 13M & 2.0G & 78.4\% \\ \bottomrule
\end{tabular}%
}
\end{table}

\vspace{2pt}
\noindent \textbf{Design principles.} Recall that our goal is to develop a ViT model suitable for deployment on IoT devices with the assistance of the cloud server.
In this section, we aim to explore how to reduce the complexity of the standard FFN layer while trading off the model performance to a small extent.
We again consider the image classification task on the ImageNet-1K dataset for ablative studies. 

On the one hand, a straightforward (or rather naive) method for improving the efficiency of an FFN is to reduce the expansion ratio $r$ (see Fig.~\ref{fig:FFNwithratio}). 
However, as we empirically show in Table~\ref{tab:abl_ffn}, reducing $r$ results in a significant degradation in model performance, e.g., the top-1 accuracy drops by almost 8\% as we reduce $r$ from 4 to 1. 
Despite the appealing improvement in model efficiency, such degradation in model performance renders the approach of reducing the expansion ratio $r$ unacceptable. 

On the other hand, another method for improving model efficiency is to reduce the number of connections while keeping the number of channels intact. 
We can adjust the number of channels by adjusting the group number g.
Note that when g=1, all channels of input and output are connected.
When g$\textgreater 1$, the channels are divided into g groups, the channels within each group are fully connected, and there is no connection between different groups.
One realization of this idea is the groupwise connected layer (e.g., group convolution \cite{Xie@Aggregated}).
%
However, as pointed out by Zhang {et al.}, a standalone groupwise connected layer is equivalent to training multiple independent small networks in parallel \cite{Zhang@ShuffleNet}, resulting in low efficiency of utilizing channel information.
Inspired by~\cite{Zhang@ShuffleNet}, we shuffle the channels between the two groupwise connected layers to achieve ``cross-talk'' among groups (see Fig.~\ref{fig:proposedPCS-FFN}). 
We name this design \underline{p}artially \underline{c}onnected and \underline{s}huffled FFN (PCS-FFN). 

\vspace{2pt}
\noindent \textbf{Effectiveness.} Evidently, as suggested by the results in Table~\ref{tab:abl_ffn}, PCS-FFN leads to a substantially better efficiency-performance tradeoff than existing methods. 
Specifically, as shown, PCS-FFN with four groups (i.e., PCS-FFN($r$=4,g=4)) is 2.4\% more accurate than the standard FFN with $r$=1 (i.e., Standard FNN ($r$=1)), while being equivalent in both the number of parameters and FLOPs. 
As another reference for performance comparison, we also implement a Ghost FFN based on the concept (i.e., using depthwise operations to partially replace dense operations) proposed in \cite{Han@GhostNet}.  
Given the empirical evidences provided in Table~\ref{tab:abl_ffn}, we set the hyperparameters of PCS-FFN to 4 for expansion ratio $r$ and 2 for number of groups $g$ for all experiments in the remainder of this paper.

\section{Experiments} \label{ref-experiments}
In this section, we first introduce our experimental setup, including the datasets, baselines, implementation details, and evaluation metrics used in this work. 
We then provide experimental comparisons with various state-of-the-art methods for image classification, object detection, and semantic segmentation tasks.

\subsection{Experimental Setup}
\subsubsection{Datasets}

\begin{table}[t]
\centering
\caption{Benchmark datasets for evaluation.\label{tab:dataset_summary}}
\resizebox{.48\textwidth}{!}{%
\begin{tabular}{@{\hspace{2mm}}lcccc@{\hspace{2mm}}}
\toprule
Dataset & Task & Train Size & Valid Size & Image Size \\ \midrule
ImageNet-1K~\cite{Russakovsky@ImageNet} & \begin{tabular}[c]{@{}c@{}}Image\\ classification\end{tabular} & 1.28M & 50K & 224$\times$224 \\
MS COCO~\cite{Lin@COCO} & \begin{tabular}[c]{@{}c@{}}Object\\ detection\end{tabular} & 118K & 5K & 1280$\times$800 \\
ADE20K~\cite{Zhou@Scene} & \begin{tabular}[c]{@{}c@{}}Semantic\\ segmentation\end{tabular} & 20K & 2K & 512$\times$512 \\ \bottomrule
\end{tabular}%
}
\end{table}

We conduct experiments on three large-scale and challenging benchmark datasets, i.e., ImageNet-1K~\cite{Russakovsky@ImageNet}, MS COCO~\cite{Lin@COCO}, and ADE20K~\cite{Zhou@Scene}, for image classification, object detection, and semantic segmentation tasks, respectively. See Table~\ref{tab:dataset_summary} for an overview.

\textbf{ImageNet} is one of the cornerstone datasets for quantifying progression in computer vision. ImageNet-1K is a subset of ImageNet, which consists of 1.28 million training images and 50K validation images from 1K different classes. 

The \textbf{MS COCO} dataset comprises over 100K densely annotated images of diverse objects from 80 categories. We use the official \emph{train2017} split for training and compare detection performance on the official \emph{val2017} split.

The \textbf{ADE20K} dataset is a challenging densely annotated dataset for scene understanding. It contains 150 fine-grained semantic categories with 20K, 2K, and 3K images for training, validation, and testing, respectively. 

To be fair and consistent with prior works, we train on the training set and report results on the validation set for performance comparison.

\subsubsection{Baselines}
To verify the effectiveness of our proposed TFormer, we consider a wide range of state-of-the-art baselines, as follows:

{\bf{ResNet}}~\cite{He@Deep} is a family of classic convolutional neural networks that have been shown to be effective across a broad spectrum of vision tasks. 

{\bf{DeiT}}~\cite{Touvron@Training}, on the one hand, is the same as the original vision transformers (ViTs) \cite{Dosovitskiy@An} from the network architecture perspective. On the other hand, DeiT proposes a data-efficient training method that enables ViTs to achieve competitive performance without pretraining on millions of images.  

{\bf{PVT}}~\cite{wang2021pyramid} is a multiscale ViT as opposed to the original ViT, which is single-scaled. PVT uses a progressive shrinking pyramid to gradually reduce the spatial resolution of features, and applies spatial reduction before attention to save computations. 

{\bf{Swin-Mixer-T}}~\cite{Liu@Swin} uses local but shifted window attention to approximate full attention in ViTs. Empirically, it is shown to be as effective as full attention but with a considerable amount of savings in model complexity. 

{\bf{ResMLP}}~\cite{touvron2021resmlp} is a vision transformer-type architecture built completely upon multilayer perceptrons. 

{\bf{PoolFormer}}~\cite{yu2021metaformer} demonstrates the importance of the skeleton structure of a vision transformer
architecture, where it shows that competitive performance can be achieved with a simple average pooling operation.

\subsubsection{Evaluation Metrics}
For image classification on ImageNet-1K, we consider the standard top-1 and top-5 accuracy to compare performance. For object detection on MS COCO, we use the mean  average precision (AP) computed over multiple intersection over union (IoU) values as the primary metric to compare performance. Additionally, we also report performance under a single IoU value (i.e., AP$_{50}$, AP$_{75}$) and for small, medium, and large objects (i.e., AP$_{\mbox{\tiny S}}$, AP$_{\mbox{\tiny M}}$, and AP$_{\mbox{\tiny L}}$) separately. For semantic segmentation on ADE20K, we compute the IoU for each semantic category and then compare the mean IoU (mIoU) averaged over all categories. 
\begin{figure}[ht]
\centering
\includegraphics[width=\linewidth]{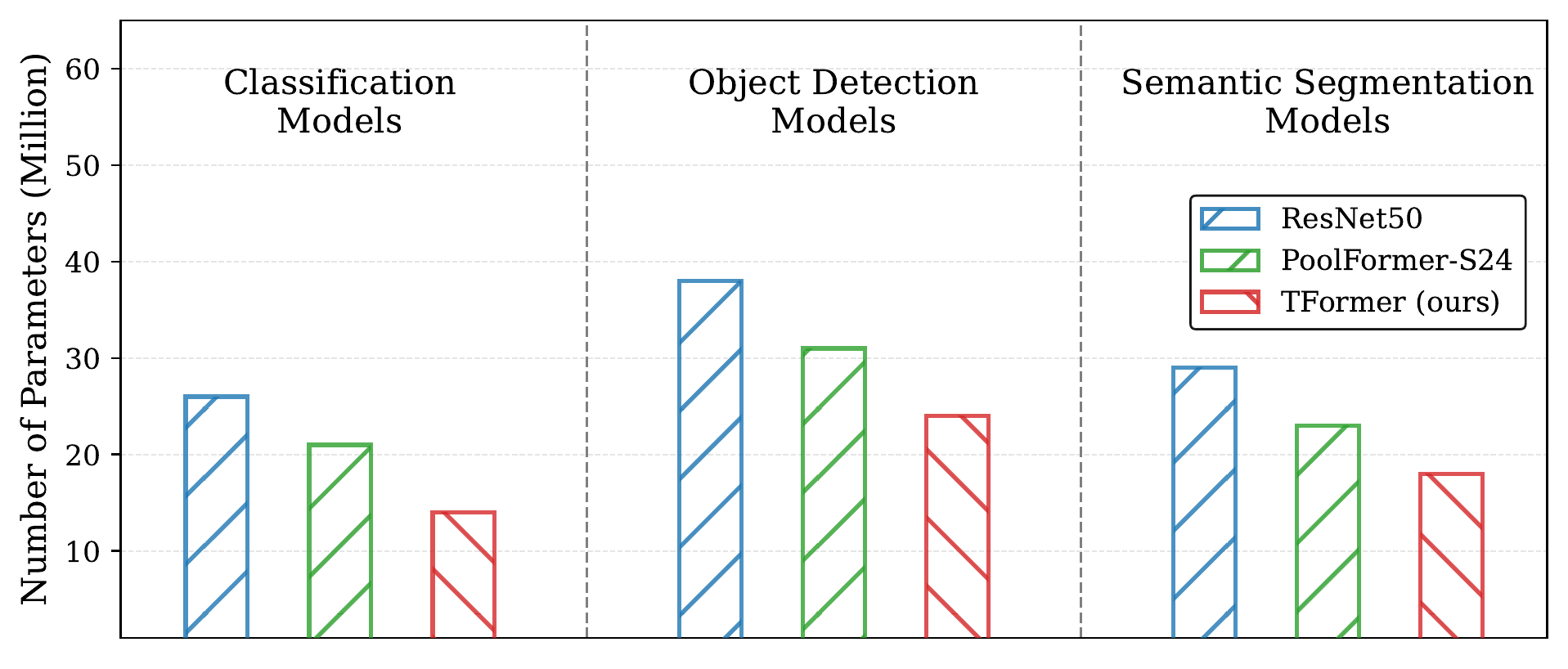}
\caption {Comparison of the number of parameters for different models on three vision tasks.}
\label{fig:network_params}
\end{figure} 
\begin{figure}[ht]
\centering
\includegraphics[width=\linewidth]{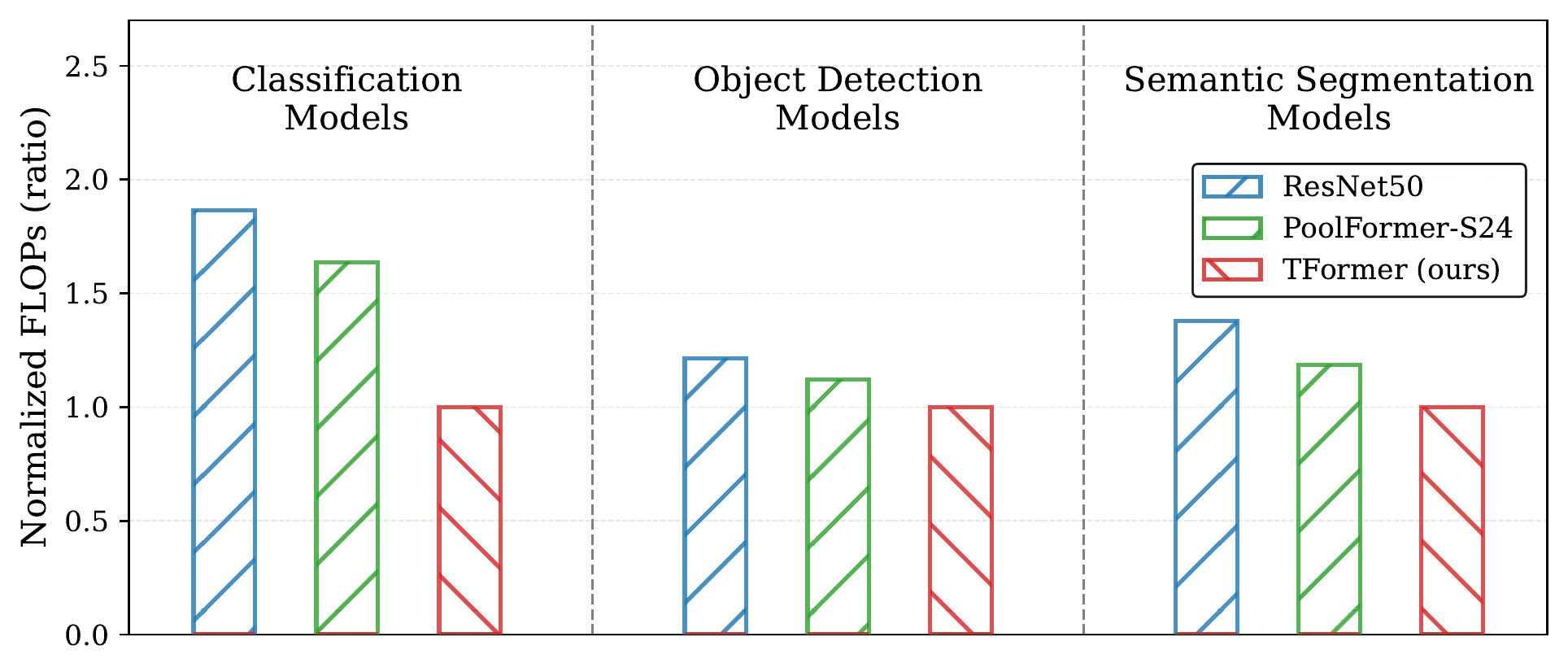}
\caption {Comparison of the normalized FLOPs for different models on three vision tasks.}
\label{fig:network_flops}
\end{figure} 

\subsubsection{Implementation Details}
We implement our method in Python 3.8 and PyTorch 1.8 with CUDA 11.1. All experiments are performed on NVIDIA 3090 GPUs. Our training setup for ImageNet-1K largely follows \cite{Touvron@Training}, where we use a combination of MixUp \cite{zhang2018mixup}, CutMix \cite{yun2019cutmix}, Cutout \cite{zhong2020random}, and RandAugment \cite{cubuk2020randaugment} for data augmentation; AdamW optimizer \cite{loshchilov2018decoupled} with weight decay 0.05 and an initial learning rate of 1e-3 for 300 epochs with a batch size of 1,024. For MS COCO, we use RetinaNet \cite{lin2017focal} as the detector. Following standard practice, we initialize the model with ImageNet-1K pretrained weights; we use the AdamW optimizer with an initial learning rate of 1e-4 and a batch size of 16 for 12 epochs. For ADE20K, we use the Semantic FPN \cite{kirillov2019panoptic} head and AdamW optimizer with a batch size of 32 for 40K minibatch iterations. We use an initial learning rate of 2e-4 with the polynomial decay schedule with a power of 0.9. It is worth noting that we adopt the advance training recipe introduced in \cite{wightman2021resnet} for training ResNet models to ensure a fair comparison. The codes will be made publicly available.
%
\begin{figure*}[ht]
    \centering
    \begin{subfigure}[b]{0.325\textwidth}
    \centering
    \includegraphics[width=\textwidth]{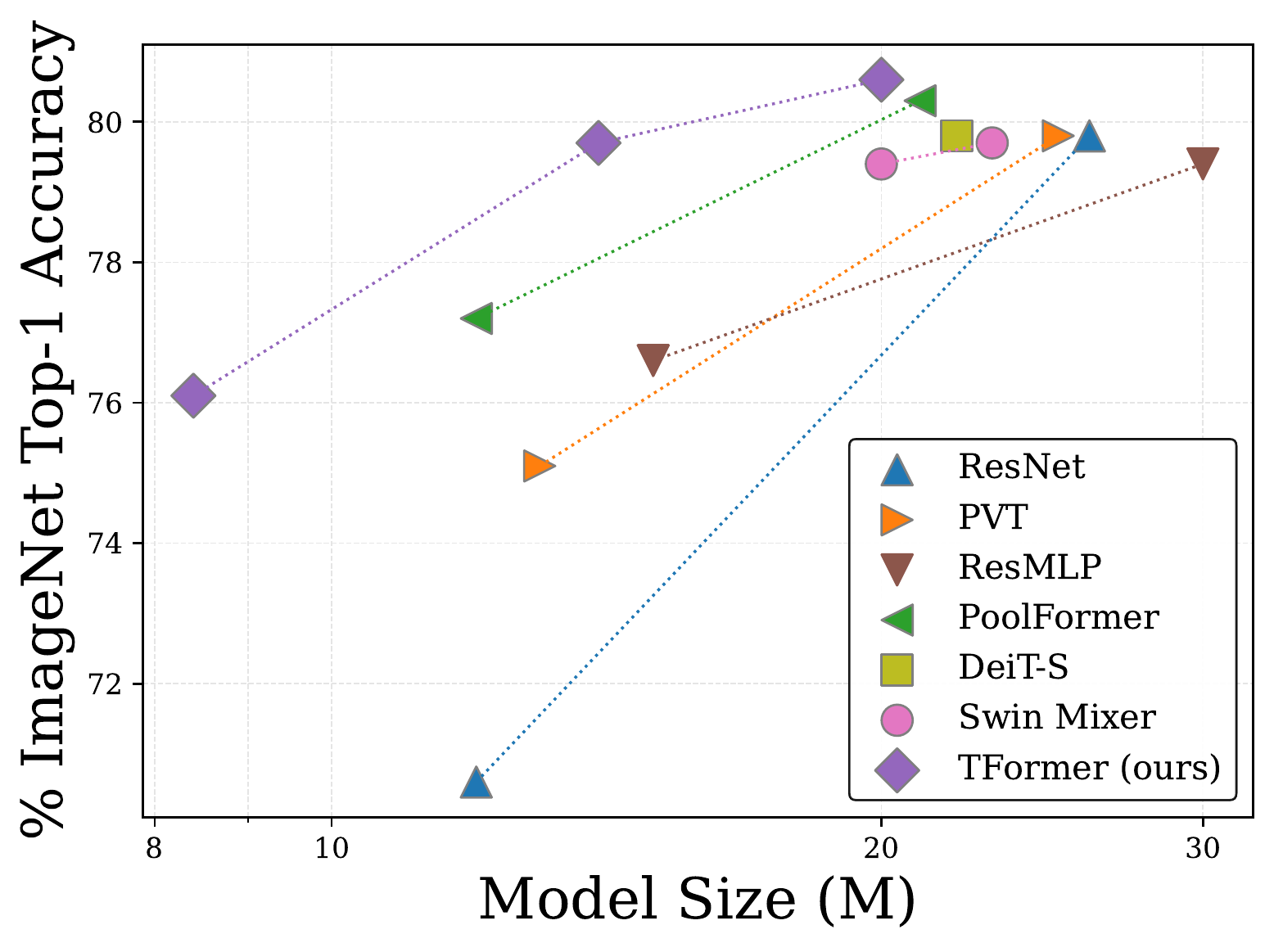}
    \caption{ImageNet-1K classification \label{fig:imagenet}}
    \end{subfigure}\hfill
    \centering
    \begin{subfigure}[b]{0.325\textwidth}
    \centering
    \includegraphics[width=\textwidth]{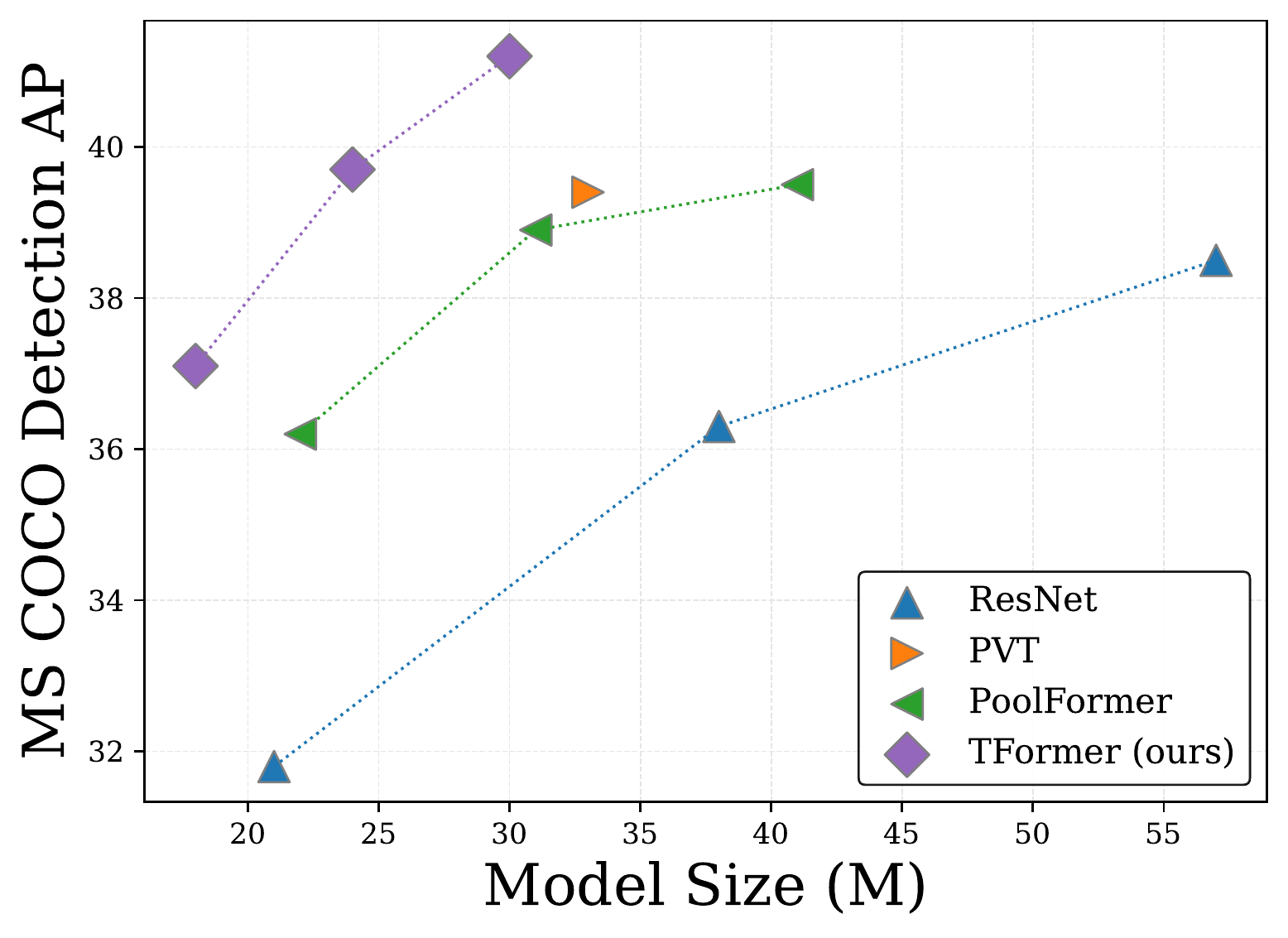}
    \caption{MS COCO object detection \label{fig:imagenet}}
    \end{subfigure}\hfill
    \centering
    \begin{subfigure}[b]{0.325\textwidth}
    \centering
    \includegraphics[width=\textwidth]{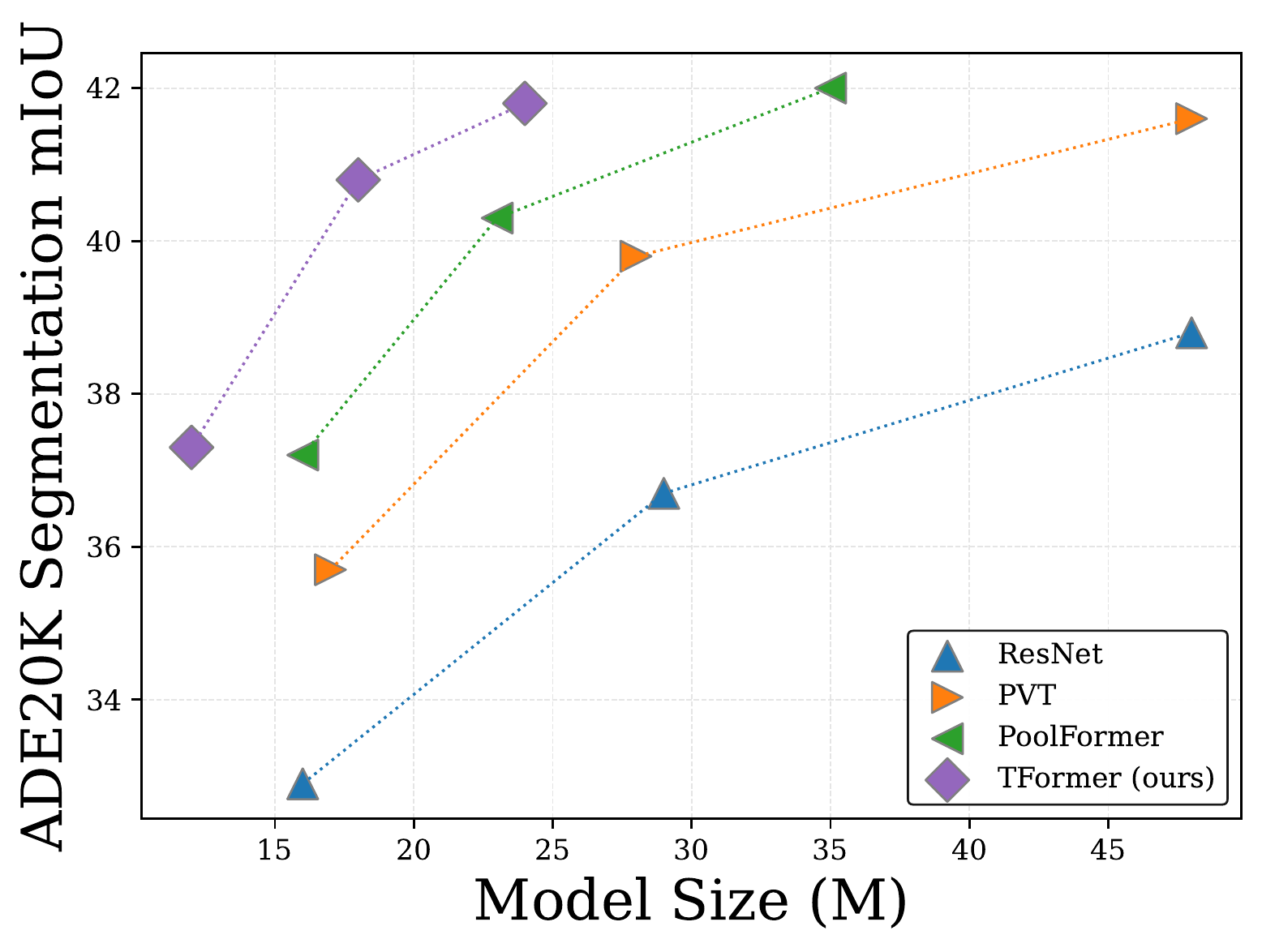}
    \caption{ADE20K semantic segmentation \label{fig:imagenet}}
    \end{subfigure}
    \caption{Accuracy and model size on different tasks of our \ourmethod{} compared to the state-of-the-art.\label{fig:result_overview}}
\end{figure*}
%
\begin{table*}[t]
\centering
\caption{Image classification performance on ImageNet-1K \cite{Russakovsky@ImageNet}. Savings (ratio) measure the parameter or FLOPs ratios between compared approaches and our \ourmethod{}s. \label{tab:in-1k}}
\resizebox{.9\textwidth}{!}{%
\begin{tabular}{@{\hspace{2mm}}l|cc|cc|cc@{\hspace{2mm}}}
\toprule
Model & \#Params (M) & Savings (ratio) & \#MAdds (G) & Savings (ratio) & Top-1 Acc (\%) & Top-5 Acc (\%) \\ \midrule
ResNet18 \cite{He@Deep} & 12 & 1.4$\times$ & 1.8 & 1.5$\times$ & 70.6 & 89.6 \\
PVT-Tiny \cite{wang2021pyramid} & 13 & 1.5$\times$ & 1.9 & 1.6$\times$ & 75.1 & 92.4\\
ResMLP-S12 \cite{touvron2021resmlp} & 15 & 1.8$\times$ & 3.0 & 2.5$\times$ & 76.6 &  93.2 \\
\ourmethod{}-S (ours) & \textbf{8.4} & \textbf{1.0$\times$} & \textbf{1.2} & \textbf{1.0$\times$} & 76.1 & 92.9 \\ \midrule
DeiT-S (0.3-12) \cite{yu2022width} & 15 & 1.1$\times$ & 3.1 & 1.4$\times$ & 78.6 & 94.4 \\
Swin-Mixer-T/D24 \cite{Liu@Swin} & 20 & 1.4$\times$ & 4.0 & 1.8$\times$ & 79.4 & 94.6 \\
ResNet50 \cite{He@Deep} & 26 & 1.9$\times$ & 4.1 & 1.9$\times$ & 79.8 & 94.5\\
DeiT-S \cite{Touvron@Training} & 22 & 1.6$\times$ & 4.6 & 2.1$\times$ & 79.8 & 95.0\\
PVT-Small \cite{wang2021pyramid} & 25 & 1.8$\times$ & 3.8 & 1.7$\times$ & 79.8 & 95.0 \\
\ourmethod{}-M (ours) & \textbf{14} & \textbf{1.0$\times$} & \textbf{2.2} & \textbf{1.0$\times$} & 79.7 & 94.8 \\ \midrule
ResMLP-S24 & 30 & 1.5$\times$ & 6.0 & 1.9$\times$ & 79.4 & 79.4 \\
Swin-Mixer-T/D6 & 23 & 1.2$\times$ & 4.0 & 1.3$\times$ & 79.7 & 94.9 \\
PoolFormer-S24 & 21 & 1.1$\times$ & 3.6 & 1.1$\times$ & 80.3 & 95.0 \\
\ourmethod{}-L (ours) & \textbf{20} & \textbf{1.0$\times$} & \textbf{3.2} & \textbf{1.0$\times$} & \textbf{80.6} & \textbf{95.4} \\ 
\bottomrule
\end{tabular}%
}
\end{table*}

\begin{table*}[ht]
\centering
\caption{Object detection performance on MS COCO \cite{Lin@COCO}.\label{tab:mscoco}}
\resizebox{.65\textwidth}{!}{%
\begin{tabular}{@{\hspace{2mm}}lccccccc@{\hspace{2mm}}}
\toprule
Model & \#Params (M) & AP & AP$_{50}$ & AP$_{75}$ & AP$_{\mbox{\tiny S}}$ & AP$_{\mbox{\tiny M}}$ & AP$_{\mbox{\tiny L}}$ \\ \midrule
ResNet18 \cite{He@Deep} & 21 & 31.8 & 49.6 & 33.6 & 16.3 & 34.3 & 43.2 \\
PoolFormer-S12 \cite{yu2021metaformer} & 22 & 36.2 & 56.2 & 38.2 & 20.8 & 39.1 & 48.0 \\
\ourmethod{}-S (ours) & 18 & 37.1 & 56.9 & 39.5 & 20.8 & 40.2 & 49.7 \\ \midrule
ResNet50 \cite{He@Deep} & 38 & 36.3 & 55.3 & 38.6 & 19.3 & 40.0 & 48.8 \\
PoolFormer-S24 \cite{yu2021metaformer} & 31 & 38.9 & 59.7 & 41.3 & 23.3 & 42.1 & 51.8 \\
\ourmethod{}-M (ours) & 24 & 39.7 & 59.9 & 42.4 & 23.8 & 43.3 & 52.9 \\ \midrule
ResNet101 \cite{He@Deep} & 57 & 38.5 & 57.8 & 41.2 & 21.4 & 42.6 & 51.1 \\
PoolFormer-S36 \cite{yu2021metaformer} & 41 & 39.5 & 60.5 & 41.8 & 22.5 & 42.9 & 52.4 \\
\ourmethod{}-L (ours) & 30 & 41.2 & 61.7 & 43.9 & 24.2 & 44.5 & 55.6 \\ \bottomrule
\end{tabular}%
}
\end{table*}
%

\begin{figure*}[t]
\centering
\includegraphics[width=0.8\linewidth]{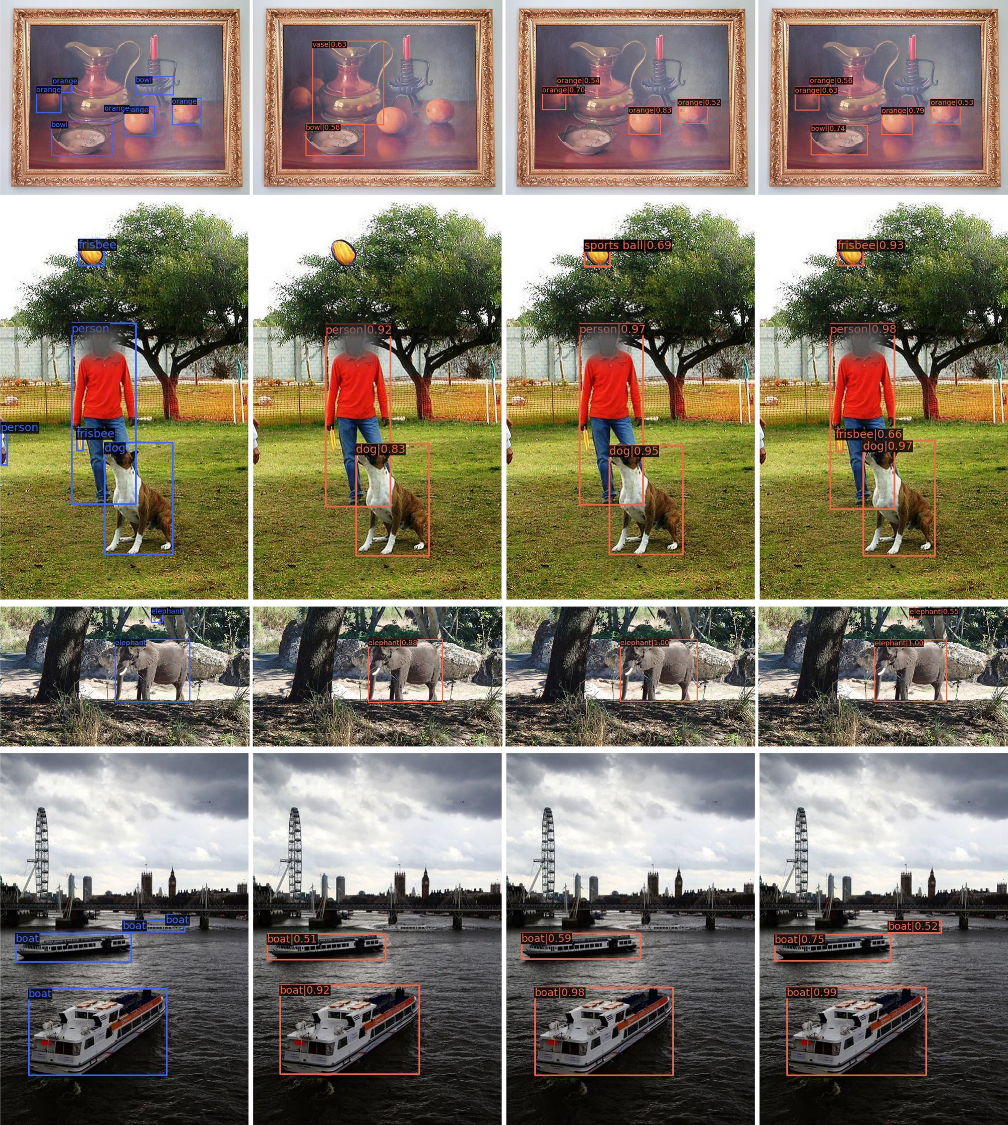}
\caption {Qualitative comparison on MS COCO dataset. From left to right, we show the example predictions from the ground truth, ResNet50, PoolFormer-S24, and TFormer-S24. The predicted labels with confidence scores are annotated at the top-left corners of the detection boxes.}
\label{fig:objectdetection}
\end{figure*} 

\begin{figure*}[t]
\centering
\includegraphics[width=.8\linewidth]{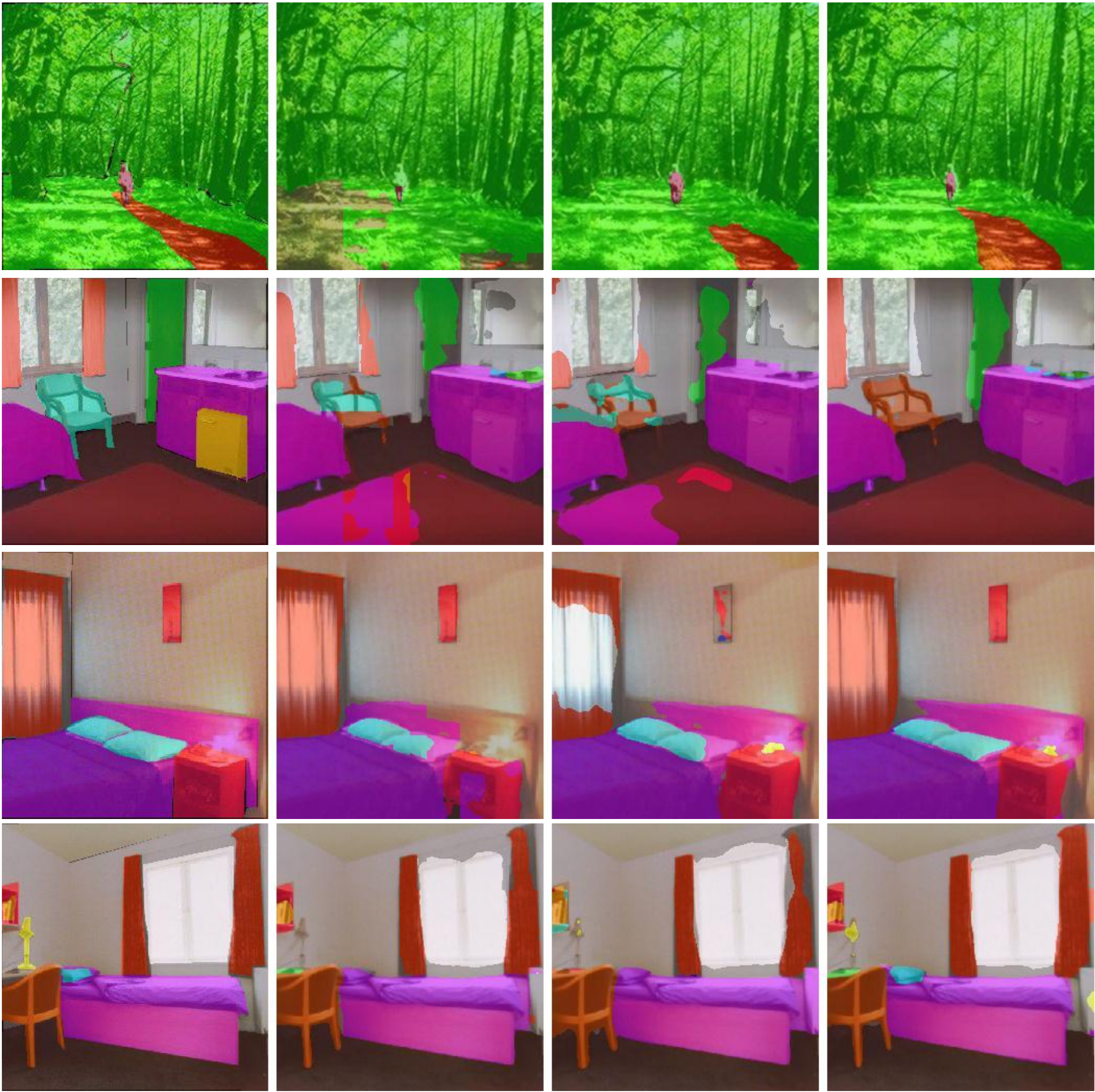}
\caption {Qualitative comparison on ADE20K dataset. We visualize ground truth, ResNet50, PoolFormer-S24, and TFormer-24 from left to right.}
\label{fig:semanticsegmentation}
\vspace{-0.2cm}
\end{figure*} 

\subsection{Experimental Results}
\subsubsection{Reduction in Model Parameter Transmission}
The number of model parameters determines the quantity of data that the cloud server needs to send to the IoT device during cloud-assisted training of \ourmethod{} and the storage resources of the IoT device that the model needs to consume.
Therefore, the number of model parameters is a measure of whether a model is suitable for cloud-assisted deployment on IoT devices.
To this end, we compare the number of parameters that \ourmethod{} has with other models when dealing with different vision tasks, as shown in Fig.~\ref{fig:network_params}.

Compared to other models, our proposed \ourmethod{} has the smallest number of model parameters in all tasks.
As shown in Table~\ref{tab:in-1k}, the number of model parameters of our \ourmethod{} is only half of the number of model parameters of ResNet50.
In addition, as shown in Table~\ref{tab:mscoco}, the amount of model parameters that need to be transmitted is 57 million when using ResNet101, and 30 million when using our proposed TFormer-L.
Compared with ResNet101, our proposed model saves 47\% of parameters and improves AP by 2.7\%.
The reasons why \ourmethod{} can reduce the number of model parameters are i) replacing the multihead attention layer with the hybrid layer containing a small number of parameters and ii) introducing the PCS-FFN module, which sparsifies the connections of neural units.
Therefore, \ourmethod{} has the smallest number of model parameters.

\subsubsection{Reduction in FLOPs}
Floating-point operations (FLOPs) are often used to measure model complexity and can represent the computing power requirements of the model for IoT devices.
To this end, to illustrate which model is more suitable for deployment on IoT devices, we compare the FLOPs of multiple models on different tasks, as shown in Fig.~\ref{fig:network_flops}.

Compared to other models, our proposed \ourmethod{} has the smallest number of FLOPs in all tasks.
As shown, taking our \ourmethod{}-M as the baseline, ResNet50 uses almost $2\times$ more FLOPs than our method on the image classification task;
meanwhile, our method also leads to $1.6\times$ and $2.1\times$ reductions in the number of parameters and FLOPs respectively when compared to the original ViT model (i.e., DeiT-S \cite{Touvron@Training}).
On object detection and semantic segmentation tasks, our \ourmethod{} also has the fewest FLOPs compared to the FLOPs of ResNet50 and PoolFormer-S24.
The reasons why \ourmethod{} can reduce the number of FLOPs are i) the introduction of the hybrid layer and ii) the introduction of the PCS-FFN module.
Therefore, \ourmethod{} has the smallest number of FLOPs and is more suitable for deployment on resource-constrained IoT devices.

\subsubsection{Performance Improvement}

{\bf{Image Classification:}}
Our proposed \ourmethod{} consistently outperforms other peer models with similar or smaller numbers of parameters in the image classification task.
As shown in Table~\ref{tab:in-1k}, 
compared with PVT-Tiny, the top-1 accuracy of \ourmethod{}-M is 4.6\% higher than that of PVT-Tiny, and the top-5 accuracy of \ourmethod{}-M is 2.4\% higher than that of PVT-Tiny;
In addition, as shown, \ourmethod{} outperforms models even with more parameters than it does.
Compared with PVT-Tiny, the top-1 accuracy of \ourmethod{}-S is 1\% higher than that of PVT-Tiny, and the top-5 accuracy of \ourmethod{}-S is 0.5\% higher than that of PVT-Tiny.
Compared with PVT-Small, the top-1 accuracy of \ourmethod{}-L is 0.8\% higher than that of PVT-Small, and the top-5 accuracy of \ourmethod{}-L is 0.4\% higher than that of PVT-Small.

\vspace{2pt}
\noindent{\bf{Object Detection:}}
Our proposed \ourmethod{} consistently outperforms other peer models with a similar or slightly larger number of parameters in the object detection task. 
As shown in Table~\ref{tab:mscoco}, \ourmethod{}-L achieves 2.3 higher AP points than PoolFormer-S24 while using a similar number of parameters. \ourmethod{}-M achieves 3.5 higher AP points than PoolFormer-S12 while using a similar number of parameters.
In addition, \ourmethod{}-S achieves 0.9 higher AP points than PoolFormer-S12 while using a smaller number of parameters. With only half the number of model parameters of ResNet101, the AP point of \ourmethod{}-L is 2.7 higher than that of ResNet101. 
In addition, we also provide a qualitative visualization between \ourmethod{}-M and the compared models in Fig.~\ref{fig:objectdetection}.

\vspace{2pt}
\noindent{\bf{Semantic Segmentation:}}
Our proposed \ourmethod{} consistently outperforms other peer models with a similar number of parameters in the semantic segmentation task. 
As shown in Table~\ref{tab:ade20k}, compared with PoolFormer-S24, our \ourmethod{}-M achieves a 0.5\% higher mIoU and a reduced number of 5M parameters, and our \ourmethod{}-L achieves a 1.5\% higher mIoU while using a similar number of parameters.
Note that compared with PoolFormer-S36, our \ourmethod{}-L reduces the number of model parameters by 11 M at the expense of 0.2\% mIoU.
A qualitative comparison is provided in Fig.~\ref{fig:semanticsegmentation}.
\begin{table}[ht]
\centering
\caption{Semantic segmentation performance on ADE20K \cite{Zhou@Scene}.\label{tab:ade20k}}
\resizebox{.4\textwidth}{!}{%
\begin{tabular}{@{\hspace{2mm}}l@{\hspace{8mm}}cc@{\hspace{2mm}}}
\toprule
Model & \#Params (M) & mIoU (\%) \\ \midrule
ResNet18 \cite{He@Deep} & 16 & 32.9 \\
PVT-Tiny \cite{wang2021pyramid} & 17 & 35.7 \\
PoolFormer-S12 \cite{yu2021metaformer} & 16 & 37.2 \\
\ourmethod{}-S (ours) & 12 & 37.3 \\ \midrule
ResNet50 \cite{He@Deep} & 29 & 36.7 \\
PVT-Small \cite{wang2021pyramid} & 28 & 39.8 \\
PoolFormer-S24 \cite{yu2021metaformer} & 23 & 40.3 \\
\ourmethod{}-M (ours) & 18 & 40.8 \\ \midrule
ResNet-101 \cite{He@Deep} & 48 & 38.8 \\
PVT-Medium \cite{wang2021pyramid} & 48 & 41.6 \\
PoolFormer-S36 \cite{yu2021metaformer} & 35 & 42.0 \\
\ourmethod{}-L (ours) & 24 & 41.8 \\ \bottomrule
\end{tabular}%
}
\end{table}

\vspace{2pt}
\noindent\textbf{Discussion.}
Fig.~\ref{fig:result_overview} clearly shows the performance and model size of \ourmethod{} and other models on different tasks.
As shown, on image classification, object detection, and semantic segmentation tasks, the proposed \ourmethod{} consistently outperforms a wide range of existing alternatives with similar or fewer parameters. 
The main reasons are two-fold.
Firstly, it is because \ourmethod{} includes different types of pooling (i.e., max-pooling and avg-pooling) with different kernel sizes (i.e., $3\times 3$, $5\times 5$, $7\times 7$, $9\times 9$, $11\times 11$).
Different types of pooling can extract multiple types of features, and pooling of different kernel sizes can extract multiple-scale features.
From~\cite{Vaswani@Attention}, we observe that, multitype and multiscale features contribute to the high performance of \ourmethod{}.
In addition, the introduction of the pooling operation and channel splitting greatly reduces the number of model parameters in \ourmethod{}.
Secondly, it is because of the introduction of group convolution and shuffle channel techniques.
The group convolution greatly reduces the number of model parameters, and the shuffle channel enables the features of different groups to flow fully.
Therefore, \ourmethod{} proposed in this paper improves the performance of the model while reducing the number of model parameters, which is of great significance in the context of the Internet of Everything for applications that use the cloud server to assist IoT device deployment and update vision transformer models.

In addition, three model variants give more options for IoT devices with different resource configurations. 
For example, for smart cameras with very limited computing and storage resources, \ourmethod{}-S can be preferentially configured, and for mobile phones with relatively sufficient resources, \ourmethod{}-L can be preferentially selected.
\section{Conclusion and Future Work\label{ref-conclusion}}
This paper presented a transmission-friendly vision transformer, namely, \ourmethod{}, for IoT devices.
In \ourmethod{}, a hybrid layer consisting of a nonlearnable layer and a pointwise convolution and a partially connected and shuffled feedforward network (PCS-FFN) consisting of group convolution and channel shuffle techniques is introduced to reduce the number of parameters and floating-point operations (FLOPs).
In addition, the proposed hybrid layer can extract multitype and multiscale features of the data, enabling \ourmethod{} to achieve high performance.
Experimental results show that \ourmethod{} can effectively improve the performance of the model on multiple tasks while reducing the number of model parameters and FLOPs.

In future work, we plan to deploy \ourmethod{} on IoT devices (e.g., Raspberry Pi 4B). 
Because IoT devices have the characteristics of dynamically changing available resources, to provide uninterrupted services, it is necessary to deploy multiple \ourmethod{}s of different capacities.
However, deploying multiple \ourmethod{}s is constrained by the limited resources of the IoT device.
To this end, we are going to study deploying multiple \ourmethod{}s in a model parameter sharing method to save IoT device storage resources.
Moreover, in addition to reducing the number of parameters, we will also explore ways to reduce the amount of data and better processing methods.

\section*{Acknowledgements}
C. Ding was supported by the Fundamental Research Funds for the Central Universities (2021RC272), the National Natural Science Foundation of China (62202039), and the China Postdoctoral Science Foundation (2021M700364); 
Z. Lu was supported by the National Natural Science Foundation of China (62106097) and the China Postdoctoral Science Foundation (2021M691424); 
S. Wang was supported by the National Natural Science Foundation of China (61922017).

\bibliographystyle{IEEEtran}
\bibliography{IEEEexample}

\end{document}